\documentclass{article}

\usepackage[preprint]{report}

\usepackage[utf8]{inputenc} 
\usepackage[T1]{fontenc}    
\usepackage{hyperref}       
\usepackage{url}            
\usepackage{booktabs}       
\usepackage{amsfonts}       
\usepackage{nicefrac}       
\usepackage{xcolor}         

\usepackage{amsmath}
\usepackage{multirow}
\usepackage{graphicx}
\usepackage{arydshln}

\usepackage{makecell}

\usepackage{longtable}
\usepackage{geometry}
\usepackage{subcaption}
\usepackage{threeparttable}

\def\equationautorefname#1#2\null{
	Eq. (#2\null)
}

\title{OmniVoice: Towards Omnilingual Zero-Shot Text-to-Speech with Diffusion Language Models}

\author{%
  \textbf{Han Zhu, Lingxuan Ye, Wei Kang, Zengwei Yao, Liyong Guo, Fangjun Kuang} \\
  \textbf{Zhifeng Han, Weiji Zhuang, Long Lin, Daniel Povey} \\
  Xiaomi Corp., China\\
  \texttt{\{zhuhan3,dpovey\}@xiaomi.com} \\
}

\begin{document}

\maketitle

\begin{abstract}
We present OmniVoice, a massively multilingual zero-shot text-to-speech (TTS) model that scales to over 600 languages. At its core is a novel diffusion language model-style discrete non-autoregressive (NAR) architecture. Unlike conventional discrete NAR models that suffer from performance bottlenecks in complex two-stage (text-to-semantic-to-acoustic) pipelines, OmniVoice directly maps text to multi-codebook acoustic tokens. This simplified approach is facilitated by two key technical innovations: (1) a full-codebook random masking strategy for efficient training, and (2) initialization from a pre-trained LLM to ensure superior intelligibility. By leveraging a 581k-hour multilingual dataset curated entirely from open-source data, OmniVoice achieves the broadest language coverage to date and delivers state-of-the-art performance across Chinese, English, and diverse multilingual benchmarks. Our code and pre-trained models are publicly available\footnote{\url{https://github.com/k2-fsa/OmniVoice}}.

\end{abstract}

\section{Introduction}

Zero-shot text-to-speech (TTS) models trained on large-scale datasets have demonstrated a remarkable ability to generate high-quality speech conditioned on only a few seconds of reference audio~\cite{chen2025neural,anastassiou2024seed}. Despite these advances, most existing models support only a limited set of languages, often leaving hundreds of low-resource languages behind. Expanding language coverage is not merely a technical challenge but also a crucial step toward extending speech technologies to languages across the globe. To address this gap, we aim to develop a massively multilingual zero-shot TTS model supporting hundreds of languages. Realizing such extensive scaling requires a TTS architecture with exceptional capacity and robustness to handle diverse linguistic patterns.

In the pursuit of such scalable and high-quality TTS, current research primarily adheres to two paradigms: autoregressive (AR)~\cite{du2024cosyvoice2,guo2024fireredtts,zhou2025indextts2,jia2025ditar,wang2025spark,ye2025llasa,song2025distar,zhou2025voxcpm,cui2025glm} and non-autoregressive (NAR) models~\cite{le2023voicebox,eskimez2024e2,chen2024f5}. 
NAR models offer advantages in both inference speed (via parallel decoding) and robustness (owing to bidirectional context)~\cite{le2023voicebox,chen2024f5,yang2025pseudo}.
Within NAR frameworks, models can be broadly categorized into continuous-latent-based~\cite{chen2024f5,zhu2025zipvoice,zhu2025zipvoicedialog,jiang2025megatts} and discrete-token-based~\cite{wang2025maskgct}. The latter was shown to yield superior prosodic diversity~\cite{yang2025measuring}.

However, state-of-the-art (SOTA) discrete-token NAR systems~\cite{wang2025maskgct} typically rely on complex two-stage cascaded pipelines (text-to-semantic followed by semantic-to-acoustic). While such decoupling simplifies the training of individual modules, it also introduces significant drawbacks: (1) error propagation, where inaccuracies in semantic prediction degrade the final audio quality; and (2) information bottlenecks, where low-bitrate semantic representations sacrifice fine-grained acoustic details. While single-stage alternatives~\cite{gallego2025single} attempt to bypass these issues, they have historically lagged behind two-stage systems in terms of speech intelligibility.

To bridge this gap, we introduce OmniVoice, an architecturally streamlined yet highly effective discrete NAR TTS framework. OmniVoice employs a discrete masked diffusion objective~\cite{sahoo2024simple} with a bidirectional Transformer~\cite{vaswani2017attention} to directly map text to multi-codebook acoustic tokens, thereby bypassing the complexity and limitations of cascaded pipelines. Its core modeling philosophy extends the success of diffusion language models~\cite{nie2025large,ye2025dream} to the speech domain. We demonstrate that the potential of this minimalist architecture can be fully unleashed through two technical innovations:

(1) Full-Codebook Random Masking: Conventional multi-codebook acoustic prediction methods adopt "per-layer" masking schedules~\cite{borsos2023soundstorm,wang2025maskgct}, suffering from inefficient convergence. We propose a fully stochastic masking strategy across all codebooks that significantly enhances training efficiency and generative quality.

(2) LLM Initialization: To resolve the intelligibility issues in single-stage discrete NAR models, we initialize our backbone with pre-trained AR LLM weights to inherit linguistic knowledge, making OmniVoice the first NAR TTS model to successfully benefit from LLM initialization~\cite{du2024cosyvoice2}.

The architectural advantage of OmniVoice makes it uniquely suited for scaling across diverse linguistic contexts. 
We curated a 581k-hour multilingual dataset encompassing more than 600 languages, derived exclusively from open-source resources. Training on this dataset allows OmniVoice to achieve the most extensive language coverage reported to date, bridging the gap for hundreds of previously under-served low-resource languages.
 
Beyond its extensive language coverage, OmniVoice supports multi-dimensional controllability, including prompt denoising~\cite{zhang2025advanced,wang2024investigation}, speaker attribute-based voice design~\cite{hu2026voicesculptor}, and fine-grained paralinguistic and phonetic control~\cite{liao2025nvspeech,deng2025indextts}. These features significantly enhance its versatility, making OmniVoice well-suited for a wide range of real-world applications.

Comprehensive experimental evaluations across Chinese, English, and massively multilingual benchmarks (covering up to 102 languages) validate that OmniVoice delivers SOTA performance in intelligibility, speaker similarity and naturalness, setting a new frontier for high-quality, high-coverage multilingual TTS.

\section{Proposed Method}

This section elaborates the design of OmniVoice. We first introduce the streamlined single-stage architecture that serves as the high-performance backbone for massively multilingual modeling, followed by two core technical optimizations to boost model training efficiency and speech intelligibility. On this basis, we further present the large-scale multilingual data construction and balancing strategies to realize extensive language coverage, and finally equip the model with multi-dimensional controllability for practical deployment.

\subsection{Architecture}

\begin{figure}[t!]
	\centering
	\includegraphics[width=0.65\columnwidth]{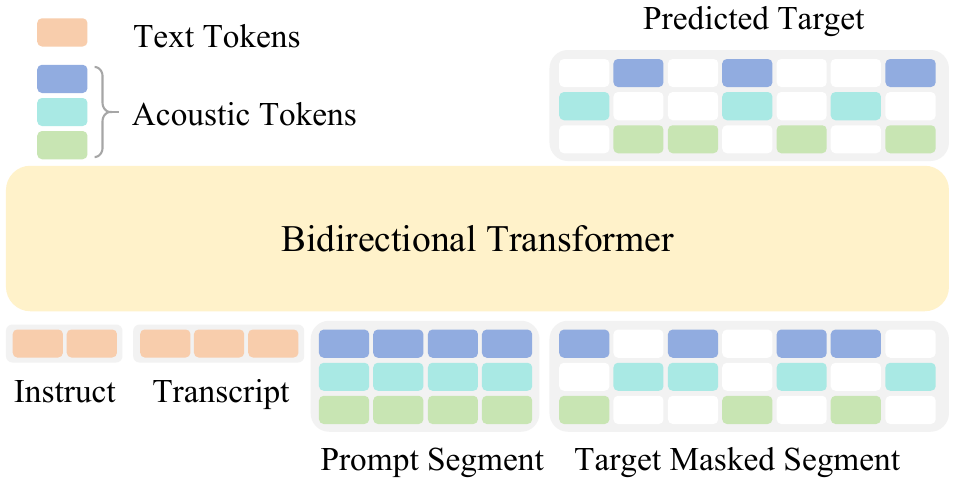}
	\caption{Illustration of OmniVoice architecture.} 
	\label{fig:architecture}
\end{figure} 

OmniVoice is a single-stage NAR TTS model with a diffusion language model-style architecture~\cite{nie2025large,ye2025dream}. Specifically, the model is trained with discrete diffusion objective and adopts a bidirectional Transformer backbone. OmniVoice directly maps text to multi-codebook acoustic tokens, eliminating the error propagation and information bottlenecks issues in conventional two-stage cascaded pipelines.
This end-to-end streamlined architecture lays the foundation for large-scale multilingual modeling, with targeted optimizations and scaling strategies detailed in the subsequent sections.

The architecture of OmniVoice is illustrated in \autoref{fig:architecture}. The input of OmniVoice comprises: 
\begin{itemize}
    \item Text token sequence ($Y$): A concatenated sequence of instruct and transcript tokens, providing the linguistic and task-oriented guidance.
    \item Acoustic token matrix ($X$): A multi-codebook matrix $X \in \mathbb{R}^{T \times C}$, where $T$ is the number of time steps and $C$ is the number of codebooks.
\end{itemize}

The acoustic matrix $X$ is partitioned along the temporal dimension into two segments: the prompt segment $X_\mathrm{prompt}$, which contains the prefix acoustic context, and the target masked segment $X_\mathrm{target}$, where tokens are randomly replaced with a special mask token $[M]$. The model is designed to leverage the text conditions $Y$, the prompt $X_\mathrm{prompt}$ and unmasked tokens in $X_\mathrm{target}$ to recover the original tokens in the masked position in $X_\mathrm{target}$.

The text tokens are embedded via a text embedding layer, and the acoustic tokens are embedded via codebook-specific embedding layers. The embeddings of all C codebooks at the same temporal position are summed to form a unified embedding, which is then fed into a bidirectional Transformer. On the output side, to reconstruct the multi-codebook tokens, OmniVoice employs $C$ independent, codebook-specific prediction heads. Each head projects the final hidden states to output a probability distribution over the vocabulary of its corresponding codebook.

The training loss is computed on the masked acoustic token positions, aiming to optimize the model for accurate token recovery. Let $\mathcal{M}$ denote the set of indices $(t, c)$ corresponding to masked positions within the target segment, where $t \in \{T_{p}+1, \dots, T\}$ and $c \in \{1, \dots, C\}$. The training loss $\mathcal{L}$ is formulated as:

\begin{equation}
\mathcal{L} = - \sum_{(t, c) \in \mathcal{M}} \log P(x_{t, c} \mid X, Y; \theta)
\end{equation}
where $x_{t, c}$ is the ground-truth acoustic token at time step $t$ and codebook index $c$, and $P(x_{t, c} \mid \dots; \theta)$ is the probability distribution predicted by the model parameterized by $\theta$.

\subsubsection{Full-Codebook Random Masking for Training Efficiency}

\begin{figure}[t!]
    \centering
	\subfloat[Per-layer masking] 
	{ \label{fig:layer_wise_mask}
		\includegraphics[width=0.35\columnwidth]{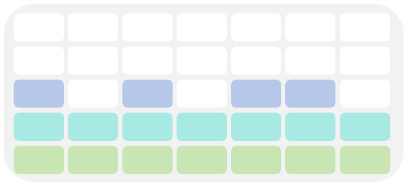}
	}
    \quad
	\subfloat[Full-codebook random masking] 
	{ \label{fig:full_codebook_random_mask}
		\includegraphics[width=0.35\columnwidth]{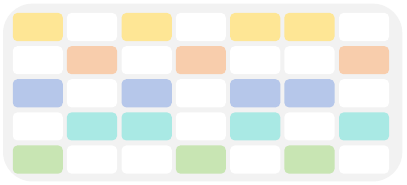}
	}
	\label{fig:decouple}
	\caption{Comparison of per-layer masking and full-codebook random masking. The x-axis denotes the time dimension, the y-axis the codebook dimension, and white blocks indicate masked tokens.} 
\end{figure}

Masking design is critical for OmniVoice, as it directly determines the training pattern and loss function.
Previous methods for multi-codebook acoustic token prediction typically employ a "per-layer" masking schedule~\cite{borsos2023soundstorm,wang2025maskgct}, which randomly masks within a single codebook layer $c$ per sample and computes the loss exclusively for that layer. Tokens in layers above the selected layer are fully masked but excluded from loss computation. Such masking strategy is designed to align with the layer-wise inference. However, it optimizes only a sparse subset of the token matrix at each iteration, leading to suboptimal training efficiency.

To circumvent this limitation, OmniVoice adopts a fully stochastic masking strategy across all codebook layers. Specifically, we independently sample a binary mask $m_{i,j} \sim \text{Bernoulli}(p_t)$ for every entry in the $T \times C$ token matrix, where the masking ratio $p_t$ is drawn from a uniform distribution $p_t \sim \mathcal{U}(0, 1)$ for each training instance. Consequently, on average, 50\% of the tokens are used for loss computation, $C$ times more than per-layer masking strategy, significantly accelerating convergence and boosting generative quality.

\subsubsection{LLM Initialization for Intelligibility}

Existing discrete NAR TTS models~\cite{wang2025maskgct,gallego2025single}, particularly single-stage architectures~\cite{gallego2025single}, often exhibit suboptimal intelligibility compared to their counterparts. To address this, we propose a simple yet effective strategy: initializing the model backbone with pre-trained AR large language models (LLMs)~\cite{yang2025qwen3}. While LLM-based initialization has been successfully integrated into recent AR-TTS frameworks~\cite{du2024cosyvoice2,du2025cosyvoice}, OmniVoice is the first NAR TTS model that successfully leverages LLM initialization to achieve notable intelligibility gains.
Our backbone is structurally identical to standard AR LLMs, facilitating direct weight transfer.
While these LLMs are originally trained with a causal mask, we empirically find that their pre-trained knowledge translates well to our bidirectional architecture.
This initialization allows OmniVoice to repurpose LLM’s linguistic capacity as a strong prior for text-to-speech mapping, significantly enhancing the intelligibility of the generated speech.

\subsection{Multilingual Scaling}

A core objective of this work is to extend language support to over 600 languages, including hundreds of low-resource languages that are rarely supported in mainstream TTS systems. OmniVoice’s streamlined end-to-end architecture is inherently well-suited for such large-scale multilingual scaling.

Expanding language coverage of TTS is a long-standing challenge~\cite{casanova2022yourtts,casanova2024xtts,zheng2025voicecraft,chen2026habibi,zhao2026lemas}. While MMS~\cite{pratap2024scaling} scales to more than 1,000 languages, yet lacks zero-shot voice cloning capabilities and relies on language-specific modeling.
Conversely, current multilingual zero-shot TTS models remain limited to narrow linguistic scopes, covering only a few dozen languages~\cite{chatterboxtts2025,liao2024fish,zheng2025voicecraft,hu2026qwen3,du2025cosyvoice,li2026indextts,zhao2026lemas}.
To bridge this gap, we introduce the first massively multilingual zero-shot TTS model capable of generalizing across hundreds of diverse languages within a single, unified model.

\begin{figure}[t!]
	\centering
	\includegraphics[width=0.8\columnwidth]{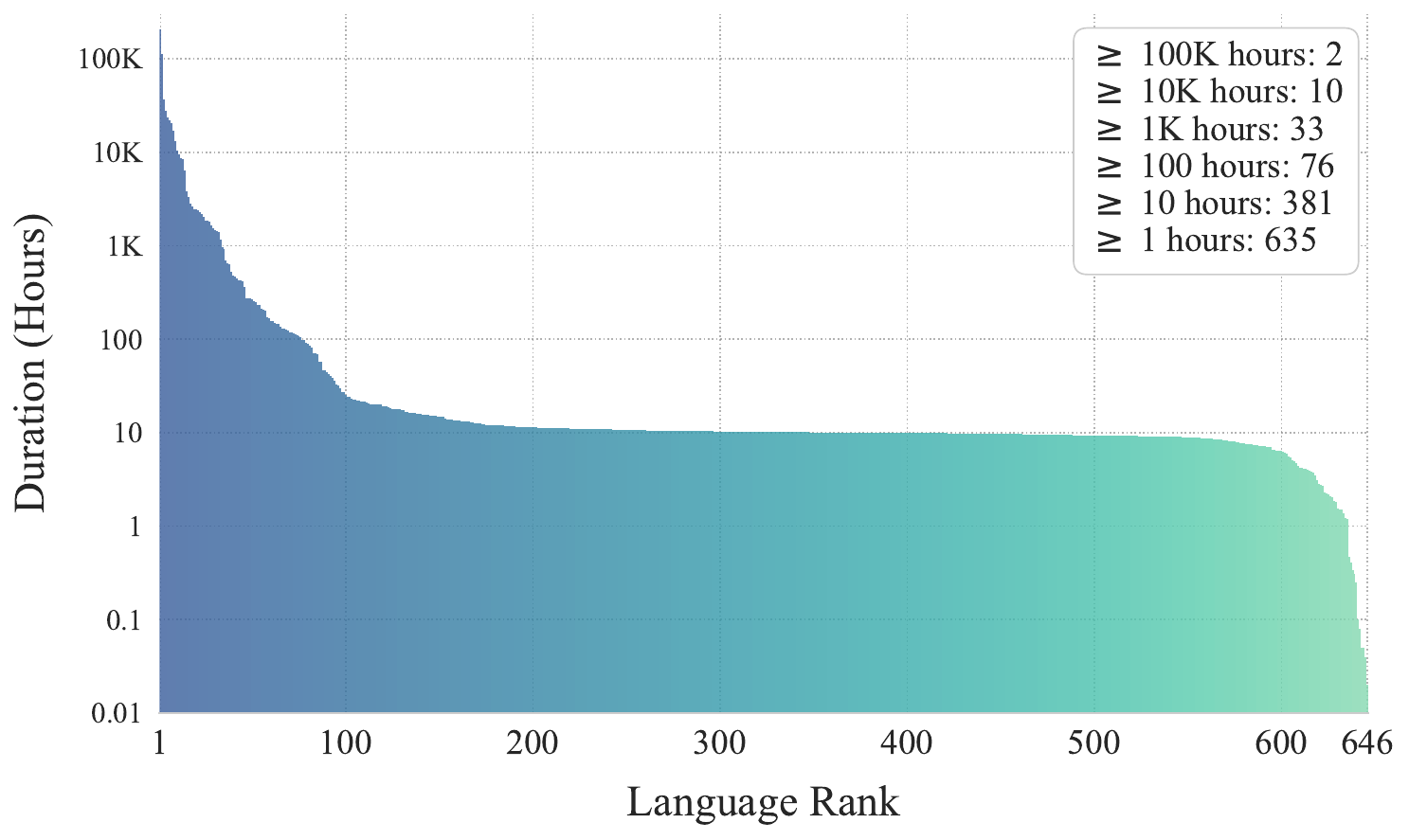}
	\caption{Statistics of the multilingual training dataset.} 
	\label{fig:language_duration}
\end{figure} 

Data acquisition remains one of the primary bottlenecks for large-scale multilingual expansion. To address this, we aggregated 50 datasets by leveraging open-source community efforts (see \autoref{sec:data_list} for the complete list). Given the heterogeneous quality of these sources (many of which were not originally designed for TTS and contain noisy audio or transcriptions), we employed a speech restoration model~\cite{nakata2025sidon} to enhance degraded speech and applied rule-based filtering~\cite{zhu2025zipvoicedialog} to exclude invalid transcriptions. The resulting corpus comprises 581k hours of audio spanning more than 600 languages (detailed statistics in \autoref{fig:language_duration}). Training on this unprecedented dataset empowers OmniVoice to achieve the broadest language coverage reported to date, significantly advancing the SOTA in multilingual TTS.

To mitigate the severe data imbalance inherent in massively multilingual datasets, where high-resource languages vastly outnumber low-resource ones, we apply a language-level data resampling strategy. Specifically, we upsample the training data of low-resource languages by assigning a repetition factor, $r_i$, to each language $i$. Let $D_i$ denote the total audio duration for language $i$, and $D_{\max}$ represent the maximum duration across all languages. The repetition factor is formulated as:

\begin{equation}
r_i = \max\left(1, \text{round}\left( \left( \frac{D_{\text{max}}}{D_i} \right)^{1 - \beta} \right) \right)
\end{equation}
where $\beta \in [0, 1]$ is a hyperparameter controlling the degree of smoothing. Setting $\beta = 1.0$ retains the natural long-tail distribution of the dataset, whereas $\beta = 0.0$ yields a uniform distribution across all languages. In our implementation, we empirically set $\beta = 0.8$ to keep the performance of high-resource languages while ensuring adequate model exposure to low-resource languages.

For multilingual text processing, we employ the subword tokenizer of pre-trained LLMs, eliminating cumbersome grapheme-to-phoneme conversion and language-specific text normalization. This strategy also allows the LLM backbone to inherit and leverage its pre-trained knowledge efficiently.

\subsection{Multi-Dimensional Controllability}

Beyond its extensive multilingual coverage, we further equip OmniVoice with multi-dimensional controllability across acoustic, identity, and linguistic aspects. This comprehensive control mechanism significantly improves the model's practicality for complex, real-world applications.

\subsubsection{Acoustic Control: Prompt Denoising}

In practical scenarios, audio prompts are often recorded in non-ideal conditions, potentially compromised by environmental noise or reverberation. To prevent the model from replicating these unwanted artifacts, we implement a prompt denoising task~\cite{zhang2025advanced,wang2024investigation}. During training, we augment a subset of the data by injecting synthetic noise and reverberation into the prompt segments. These samples are paired with a specific instruction token, \texttt{<|denoise|>}. This mechanism compels the model to disentangle the speaker's intrinsic voice identity from the acoustic environment, enabling the synthesis of clean, high-fidelity speech even when conditioned on degraded prompts.

\subsubsection{Identity Control: Speaker-Attribute-Based Voice Design}

To enable flexible TTS in the absence of an audio prompt, OmniVoice supports speaker-attribute-based voice design~\cite{hu2026voicesculptor,hu2026qwen3}. By incorporating specific speaker attributes (e.g., gender, age, pitch, and accent/dialect) into the training instruction sequence, the model can synthesize highly customized voices on demand.

\subsubsection{Linguistic Control: Paralinguistics and Phonetics}

To bridge the gap between basic intelligibility and human-like expressiveness, we incorporate paralinguistic control (e.g., laughter) by training on datasets enriched with affective cues~\cite{liao2025nvspeech,ye2025scalable}.

Furthermore, recognizing that models may occasionally encounter challenges with the pronunciation of linguistic corner cases, such as Chinese polyphonic characters or specialized English terminology, we adopt a hybrid text input format~\cite{deng2025indextts} to provide users with explicit phonetic override capabilities. During training, we stochastically replace characters or words with their corresponding phonetic transcriptions, specifically Pinyin for Chinese and phonemes from the CMU pronunciation dictionary for English. This mechanism allows for deterministic control over pronunciation during inference, enabling the model to handle complex linguistic scenarios with high precision.

\section{Experimental Setup}

This section details the training datasets, model configurations, training and inference protocols, evaluation benchmarks, and evaluation metrics of our experiments.

\subsection{Training Datasets}

We train OmniVoice under two distinct data configurations. First, a bilingual variant is trained on the Chinese and English subsets of the Emilia dataset~\cite{he2025emilia}, enabling fair comparison with existing SOTA zero-shot TTS models trained on the same data. Prompt denoising is omitted in this variant to isolate its impact and highlight the inherent advantages of our architecture.  Second, a multilingual variant is trained on a self-built 581k-hour multilingual dataset spanning 600+ languages.

\subsection{Model Details}
OmniVoice is built on a bidirectional Transformer backbone, which is initialized with the pre-trained LLM weights of Qwen3-0.6B~\cite{yang2025qwen3}. The Higgs-audio tokenizer~\cite{higgsaudio2025} is adopted to extract 8-codebook acoustic tokens and reconstruct audio from these tokens.

\subsection{Training}
The AdamW optimizer~\cite{loshchilov2017decoupled} is used, with a peak learning rate of $1e-4$ and a cosine learning rate schedule that includes 3\% of total training steps as warmup. Mixed precision (BF16) and sequence packing (8192 tokens per GPU) are employed during training to improve efficiency. Using 8 H800 GPUs, the multilingual variant (2M training updates) and the bilingual Emilia variant (300k training updates) were trained in 9.66 days and 1.33 days, respectively.

\subsection{Inference}
During inference, we perform a 32-steps iterative unmasking process. The cumulative proportion of unmasked tokens at step $n$, denoted as $r_n$, follows a time-shifted schedule that was originally used in \cite{zhu2025zipvoice}:
\begin{equation}
r_n = \frac{\tau \cdot (n/N)}{1 + (\tau - 1) \cdot (n/N)}
\end{equation}
where $N=32$ is the total number of steps, and $\tau = 0.1$ is the shift parameter. The proportion of tokens to be newly unmasked at each step $n$ is given by $k_n =  r_n - r_{n-1} $, where $r_0 = 0$. 

The selection of unmasking positions and token identities is handled as follows:
\begin{itemize}
\item \textbf{Position Selection}: At each step, we identify $k_n$ positions to be unmasked by sampling from the confidence scores in the log-softmax space. To introduce beneficial stochasticity and avoid local optima, we apply a temperature $T=5$ to the confidence scores before sampling the indices.
\item \textbf{Token Assignment}: Once the positions are selected, the specific token class for each position is determined deterministically by taking the $argmax$ of the predicted probability distribution.
\end{itemize}

We also apply a layer penalty on the confidence scores to encourage unmasking lower-layer tokens first. Furthermore, classifier-free guidance~\cite{ho2021classifierfree} is utilized in the log-softmax space with a guidance scale of 2. All aforementioned strategies are verified to effectively improve model performance and generation stability.

During inference, each character is assigned a script-dependent duration weight $w(c)$ to reflect intrinsic duration differences across writing systems (e.g., CJK, Latin). Given the prompt audio duration $D_{prompt}$, the target duration $D_{target}$ is estimated by scaling with the ratio of total character weights:
\begin{equation}
D_\mathrm{target} = D_\mathrm{prompt} \cdot \frac{W_\mathrm{target}}{W_\mathrm{prompt}}
\end{equation}
where
$W_\mathrm{target} = \sum_{c\in\mathrm{Target}} w(c)$ and $ W_\mathrm{prompt} = \sum_{c\in\mathrm{Prompt}} w(c)$.

\subsubsection{Evaluation Benchmarks}
We evaluate OmniVoice on four benchmarks covering standard Chinese/English settings and massively multilingual scenarios:
\begin{itemize}
    \item \textbf{LibriSpeech-PC}~\cite{meister2023librispeech, chen2024f5}: A standard English zero-shot TTS benchmark.
    \item \textbf{Seed-TTS}~\cite{anastassiou2024seed}: A bilingual (Chinese/English) zero-shot benchmark.
    \item \textbf{MiniMax-Multilingual-24}\cite{zhang2025minimax}: A multilingual benchmark covering 24 languages.
    \item \textbf{FLEURS-Multilingual-102}: A 102-language multilingual benchmark constructed from the dev/test splits of the FLEURS dataset~\cite{conneau2023fleurs} to further evaluate OmniVoice’s multilingual capability. It is the zero-shot TTS evaluation benchmark with the widest language coverage to date.
\end{itemize}

\subsection{Evaluation Metrics}
Our evaluation combines several objective and subjective metrics. For speaker similarity evaluation, we use SIM-o~\cite{le2023voicebox} with a WavLM-based~\cite{chen2022wavlm} ECAPA-TDNN model~\cite{desplanques2020ecapa}. Intelligibility is measured using word error rate (WER) or character error rate (CER) across different languages. 
For brevity and consistency, we refer to both WER and CER as WER for datasets using both metrics (Seed-TTS and MiniMax-Multilingual-24), while evaluating each language with its appropriate metric.
Specifically, we use the Hubert-based ASR model~\cite{hsu2021hubert} for LibriSpeech-PC test-clean, Paraformer-zh~\cite{gao2022paraformer} for Chinese, the Omnilingual ASR model~\cite{omnilingual2025omnilingual} for the FLEURS benchmark, and Whisper-large-v3~\cite{radford2023robust} for the remaining datasets. We also adopt UTMOS~\cite{saeki2022utmos} to assess objective speech naturalness. 

These objective metrics are supplemented with subjective evaluations, including comparative mean opinion score (CMOS, $[-3,3]$) and similarity mean opinion score (SMOS, $[0,5]$), which measure human opinions on relative speech quality and absolute speaker similarity to the prompt audio.

\section{Experimental Results}

\subsection{Evaluation on English and Chinese}

\begin{table}[ht]
\centering
\caption{Objective evaluation results on Chinese and English test sets. Baseline results are obtained using official checkpoints.
Params. (Parameters) denotes the total parameter size of voice cloning TTS systems (including audio tokenizer, vocoder, and other related components). 
Best results are highlighted in \textbf{bold}. 
Top: Results on LibriSpeech-PC. 
Bottom: Results on Seed-TTS test sets.}
\label{tab:objective_results}
\begin{tabular}{lccccc}
\toprule
\multirow{2}{*}{\textbf{Model}} & \multirow{2}{*}{\textbf{Params.}} & \multirow{2}{*}{\textbf{Training Data (hours)}} &  \multicolumn{3}{c}{\textbf{LibriSpeech-PC test-clean}} \\
\cmidrule(lr){4-6}
& & & \textbf{SIM-o} $\uparrow$ & \textbf{WER} $\downarrow$ & \textbf{UTMOS} $\uparrow$ \\
\midrule
\textbf{Ground-truth} & - & - & 0.690 & 1.87 & 4.10 \\
\midrule
\textit{\textbf{AR Models}} & & & & & \\
IndexTTS2 & 1.7B & 55k Multi. & 0.700 & 2.35 & 4.06 \\
CosyVoice3 & 1.1B & 1000k Multi. & 0.694 & 1.59 & 4.28 \\
VoxCPM & 0.7B & 1800k Multi. & 0.717 & 1.74 & 4.18 \\
Qwen3-TTS & 1.1B & 5000k Multi. & 0.704 & 1.60 & \textbf{4.41} \\
\midrule
\textit{\textbf{NAR Models}} & & & & & \\
F5-TTS & 0.4B & 100K Emilia & 0.655 & 1.89 & 3.89 \\
ZipVoice & 0.1B & 100k Emilia & 0.668 & 1.64 & 3.98 \\
MaskGCT & 2.2B & 100K Emilia & 0.691 & 2.26 & 3.91 \\
OmniVoice-Emilia & 0.8B & 100k Emilia & 0.697 & 1.57 & 4.23 \\
OmniVoice & 0.8B & 581k Multi. & \textbf{0.729} & \textbf{1.30} & 4.28 \\
\bottomrule
\end{tabular}
\vspace{20pt} 
\begin{tabular}{lcccccc}
\toprule
\multirow{2}{*}{\textbf{Model}} & \multicolumn{3}{c}{\textbf{Seed-TTS test-en}} & \multicolumn{3}{c}{\textbf{Seed-TTS test-zh}} \\
\cmidrule(lr){2-4} \cmidrule(lr){5-7}
& \textbf{SIM-o} $\uparrow$ & \textbf{WER} $\downarrow$ & \textbf{UTMOS} $\uparrow$ 
& \textbf{SIM-o} $\uparrow$ & \textbf{WER} $\downarrow$ & \textbf{UTMOS} $\uparrow$ \\
\midrule
\textbf{Ground-truth} & 0.734 & 2.14 & 3.52 & 0.755 & 1.25 & 2.78 \\
\midrule
\textit{\textbf{AR Models}} \\
IndexTTS2 & 0.706 & 2.33 & 3.65 & 0.764 & 1.05 & 3.00 \\
CosyVoice3 & 0.696 & 2.17 & 3.96 & \textbf{0.778} & 1.14 & 3.32 \\
VoxCPM & 0.731 & 1.92 & 3.77 & 0.772 & 0.99 & 2.94 \\
Qwen3-TTS & 0.708 & \textbf{1.54} & \textbf{4.16} & 0.766 & 1.15 & \textbf{3.46} \\
\midrule
\textit{\textbf{NAR Models}} \\
F5-TTS & 0.664 & 1.85 & 3.72 & 0.750 & 1.53 & 2.93 \\
ZipVoice & 0.697 & 1.70 & 3.82 & 0.751 & 1.40 & 3.15 \\
MaskGCT & 0.713 & 2.88 & 3.55 & 0.773 & 2.40 & 2.63 \\
OmniVoice-Emilia & 0.717 & 1.72 & 3.88 & 0.765 & 0.89 & 3.05 \\
OmniVoice & \textbf{0.741} & 1.60 & 3.91 & 0.777 & \textbf{0.84} & 3.11 \\
\bottomrule
\end{tabular}
\end{table}

\begin{table}[h!]
\caption{Subjective evaluation results on Chinese and English test sets. CMOS and SMOS are used for evaluation. Best results are highlighted in \textbf{bold}.}
\label{tab:subjective_results}
\centering
\begin{tabular}{lcc}
\toprule
\textbf{Model} & \textbf{CMOS} $\uparrow$ & \textbf{SMOS} $\uparrow$ \\
\midrule
Ground-truth & 0.00 & 3.02\scriptsize{$\pm$0.20} \\
\midrule
Qwen3-TTS & 0.40\scriptsize{$\pm$ 0.16} & 3.65 \scriptsize{$\pm$ 0.18} \\
ZipVoice & -0.30\scriptsize{$\pm$ 0.16} & 3.35\scriptsize{$\pm$ 0.19} \\
MaskGCT & -0.38\scriptsize{$\pm$ 0.17} & 3.20\scriptsize{$\pm$ 0.18} \\
OmniVoice-Emilia & 0.42\scriptsize{$\pm$ 0.15} & 3.58\scriptsize{$\pm$ 0.18} \\
OmniVoice & \textbf{0.44}\scriptsize{$\pm$ 0.16} & \textbf{3.80}\scriptsize{$\pm$ 0.17} \\
\bottomrule
\end{tabular}
\end{table}

\autoref{tab:objective_results} and \autoref{tab:subjective_results} summarize the Chinese and English performance of OmniVoice in comparison with SOTA AR/NAR TTS models.

OmniVoice-Emilia surpasses all NAR baselines (F5-TTS~\cite{chen2024f5}, ZipVoice~\cite{zhu2025zipvoice}, MaskGCT~\cite{wang2025maskgct}) trained on the same Emilia corpus, verifying the effectiveness of our proposed architecture.

The final multilingual version OmniVoice model yields competitive overall performance across all benchmarks against baselines trained on unconstrained datasets (IndexTTS2~\cite{zhou2025indextts2}, CosyVoice3~\cite{du2025cosyvoice}, VoxCPM~\cite{zhou2025voxcpm}, Qwen3-TTS~\cite{hu2026qwen3}), with particular advantages in speaker similarity and intelligibility. This demonstrates OmniVoice's strong capability on the two most high-resource languages.

\subsection{Evaluation on Multilingual Benchmarks}

We validate OmniVoice’s multilingual capability on the 24-language MiniMax-Multilingual-24 benchmark and the 102-language FLEURS-Multilingual-102 benchmark.

\begin{table}[htbp]
\centering
\resizebox{0.95\columnwidth}{!}{
\begin{threeparttable}
\caption{Evaluation on the MiniMax-Multilingual-24 test set.}
\label{tab:minimax_comparison}
\begin{tabular}{l ccc ccc}
\toprule
\multirow{2}{*}{Language} & \multicolumn{3}{c}{WER $\downarrow$} & \multicolumn{3}{c}{SIM-o $\uparrow$} \\
\cmidrule(lr){2-4} \cmidrule(lr){5-7}
& OmniVoice & MiniMax & ElevenLabs & OmniVoice & MiniMax & ElevenLabs \\
\midrule
Arabic       & \textbf{1.392} & 1.665 & 1.666  & \textbf{0.776} & 0.736 & 0.706 \\
Cantonese    & \textbf{17.709}\tnote{*}& 34.111& 51.513 & \textbf{0.838} & 0.778 & 0.670 \\
Chinese      & \textbf{1.008} & 2.252 & 16.026 & \textbf{0.821} & 0.780 & 0.677 \\
Czech        & 2.856 & 3.875 & \textbf{2.108}  & \textbf{0.837} & 0.796 & 0.685 \\
Dutch        & 1.358 & 1.143 & \textbf{0.803}  & \textbf{0.813} & 0.738 & 0.680 \\
English      & \textbf{1.560} & 2.164 & 2.339  & \textbf{0.884} & 0.756 & 0.613 \\
Finnish      & 3.750 & 4.666 & \textbf{2.964}  & \textbf{0.864} & 0.835 & 0.759 \\
French       & \textbf{3.347} & 4.099 & 5.216  & \textbf{0.801} & 0.628 & 0.535 \\
German       & 0.964 & 1.906 & \textbf{0.572}  & \textbf{0.812} & 0.733 & 0.614 \\
Greek        & 1.057 & 2.016 & \textbf{0.991}  & \textbf{0.867} & 0.826 & 0.733 \\
Hindi        & \textbf{4.330} & 6.962 & 5.827  & \textbf{0.863} & 0.818 & 0.730 \\
Indonesian   & 1.973 & 1.237 & \textbf{1.059}  & \textbf{0.805} & 0.729 & 0.660 \\
Italian      & 2.070 & \textbf{1.543} & 1.743  & \textbf{0.812} & 0.699 & 0.579 \\
Japanese     & 4.027 & \textbf{3.519} & 10.646 & \textbf{0.828} & 0.776 & 0.738 \\
Korean       & 2.651 & \textbf{1.747} & 1.865  & \textbf{0.828} & 0.776 & 0.700 \\
Polish       & 0.874 & 1.415 & \textbf{0.766}  & \textbf{0.877} & 0.802 & 0.729 \\
Portuguese   & 2.511 & 1.877 & \textbf{1.331}  & \textbf{0.859} & 0.805 & 0.711 \\
Romanian     & 2.424 & 2.878 & \textbf{1.347}  & \textbf{0.836} & 0.809 & 0.699 \\
Russian      & \textbf{2.233} & 4.281 & 3.878  & \textbf{0.783} & 0.761 & 0.676 \\
Spanish      & \textbf{1.026} & 1.029 & 1.084  & \textbf{0.804} & 0.762 & 0.615 \\
Thai         & 3.978 & \textbf{2.701} & 73.936 & \textbf{0.841} & 0.800 & 0.588 \\
Turkish      & 2.166 & 1.520 & \textbf{0.699}  & \textbf{0.851} & 0.779 & 0.596 \\
Ukrainian    & 1.774 & 1.082 & \textbf{0.997}  & \textbf{0.810} & 0.730 & 0.647 \\
Vietnamese   & 1.373 & \textbf{0.880} & 73.415 & \textbf{0.804} & 0.743 & 0.369 \\
\midrule
\textbf{Average}      & \textbf{2.850} & 3.774 & 10.950 & \textbf{0.830} & 0.766 & 0.655 \\
\bottomrule
\end{tabular}

\begin{tablenotes}
    \footnotesize
    \item[*] Cantonese WER is 2.273\% when evaluated with the SenseVoice-Small ASR model.
\end{tablenotes}
\end{threeparttable}
}
\end{table}
\normalsize

As shown in \autoref{tab:minimax_comparison}, despite being trained exclusively on open-source datasets, OmniVoice clearly outperforms leading commercial systems (ElevenLabs Multilingual v2 and MiniMax-Speech) in both average SIM-o and WER, demonstrating that OmniVoice achieves commercial-grade multilingual TTS performance. While Cantonese shows a higher WER in the table, our analysis reveals that this is due to limitations in the Whisper ASR model rather than the quality of the generated speech. When evaluated with the SenseVoice-Small ASR model~\cite{an2024funaudiollm}, the WER for Cantonese decreases to 2.273\%. However, to maintain consistency with the benchmarks in \cite{zhang2025minimax}, we retain the Whisper-based results in our report.

\begin{table}[h!]
\centering
\caption{Evaluation on the FLEURS-Multilingual-102 test set.}
\label{tab:fleurs_comparison}
\begin{tabular}{lcccc}
\toprule
\multirow{3}{*}{\textbf{Model}}
& \multicolumn{4}{c}{\textbf{FLEURS-Multilingual-102}} \\
\cmidrule(lr){2-5}
& \multirow{2}{*}{\textbf{Avg SIM-o $\uparrow$}}
& \multirow{2}{*}{\textbf{Avg CER $\downarrow$}}
& \multicolumn{2}{c}{\textbf{Languages with CER $\leq$}} \\
\cmidrule(lr){4-5}
& & & \textbf{5\%} & \textbf{10\%} \\
\midrule
Ground-truth & -     & 5.11 & 75 & 92 \\
OmniVoice    & \textbf{0.788} & \textbf{4.00} & \textbf{82} & \textbf{95} \\
\bottomrule
\end{tabular}
\end{table}

\begin{figure}[htbp]
	\centering
	\includegraphics[width=0.9\columnwidth]{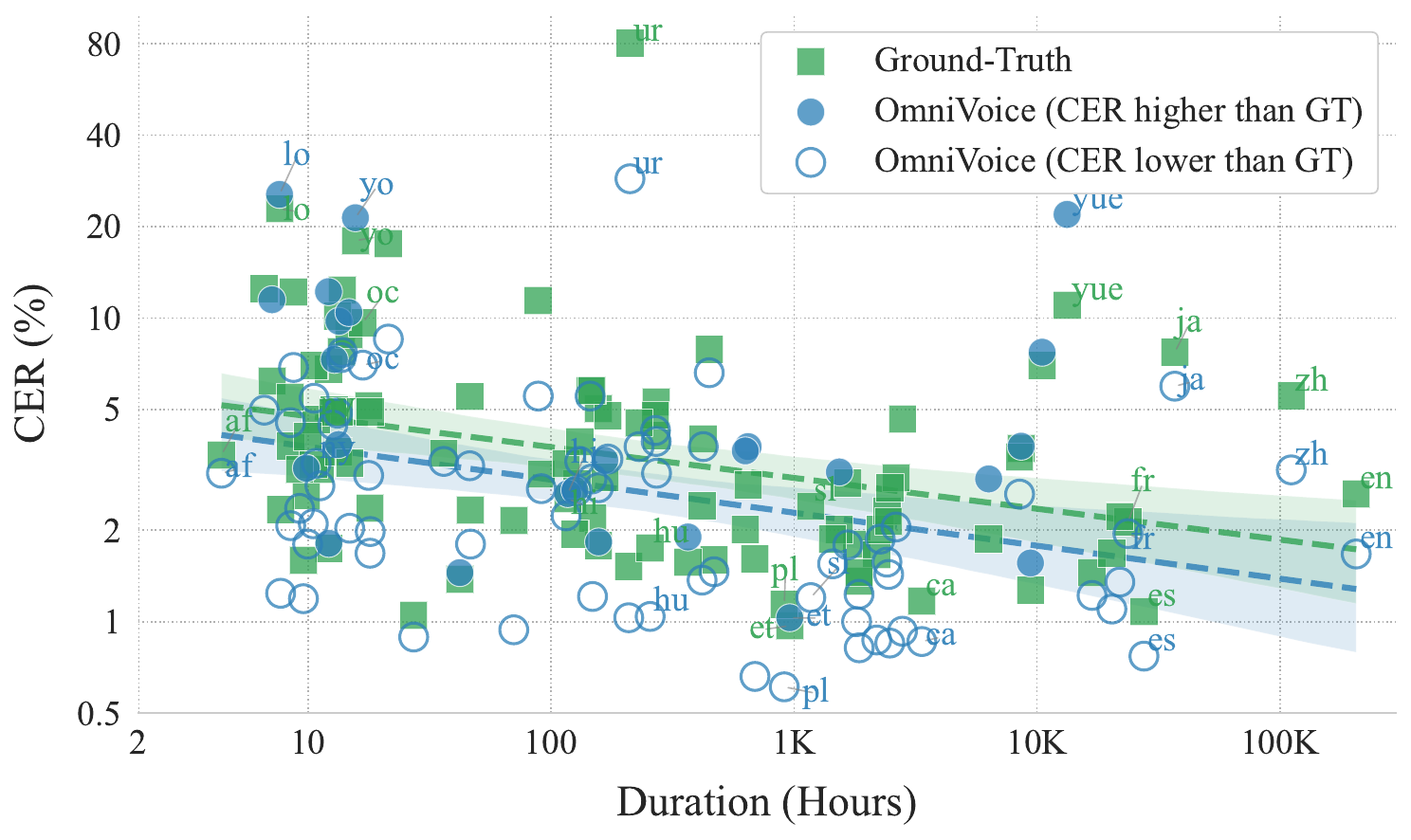}
	\caption{CERs of OmniVoice vs. ground truth across languages with varying training data durations on FLEURS-Multilingual-102. Hollow circles indicate languages where OmniVoice achieves lower CER than ground-truth, Filled circles indicate higher CER.} 
	\label{fig:fleurs_result}
\end{figure} 

Next, we evaluate OmniVoice on our self-built FLEURS-Multilingual-102 benchmark. We report the average SIM-o, average CER, and the number of languages below specific CER thresholds in \autoref{tab:fleurs_comparison}.
We observe that OmniVoice achieves an average CER of 4.00\%, which is comparable to that of the ground truth. Per-language CER results are provided in \autoref{sec:appendix_fleurs}.

To better understand OmniVoice’s performance, we plot a figure illustrating the relationship between per-language CER and training data duration in \autoref{fig:fleurs_result}.
It can be seen that OmniVoice maintains high intelligibility (CER < 5\%) even for many languages with less than 10 hours of training data, demonstrating strong generalization capability on low-resource languages. Note that we do not claim OmniVoice can generate speech of better quality than the ground truth for all languages. However, OmniVoice’s performance has exceeded the measurement capability of existing ASR models.

\subsection{Effectiveness of Key Designs}
To validate the effectiveness of our core designs, we conduct ablation experiments on the Emilia dataset. All models are trained for 300k updates with the same hyperparameters, with prompt denoising disabled unless otherwise specified.

\begin{table}[htbp]
\centering
\caption{Impact of different masking strategies.}
\label{tab:ablation_mask}
\begin{tabular}{lccc}
\toprule
\multirow{2}{*}{\textbf{Masking Strategy}} & \multicolumn{3}{c}{\textbf{Librispeech-PC test-clean}} \\
\cmidrule(lr){2-4}
& \textbf{SIM-o} $\uparrow$ & \textbf{WER} $\downarrow$ & \textbf{UTMOS} $\uparrow$ \\
\midrule
SoundStorm-style mask & 0.661 & 3.00 & 4.12 \\
MaskGCT-style mask & 0.660 & 2.04 & 4.17 \\
\midrule
Full-codebook random mask & \textbf{0.697} & \textbf{1.57} & \textbf{4.23} \\
- compute loss on a single codebook & 0.648 & 2.85 & 4.22 \\
\bottomrule
\end{tabular}
\end{table}

Firstly, we compare different masking strategies to validate the effectiveness of our proposed full-codebook random masking strategy. SoundStorm-style masking is a variant of the per-layer masking strategy, which samples layers for loss computation from a uniform distribution and selects the masking ratio according to a cosine function. MaskGCT-style masking further samples layers based on a linear distribution, where lower layers are selected with higher probability.

As shown in \autoref{tab:ablation_mask}, full-codebook random masking consistently outperforms the single-codebook masking strategies adopted by MaskGCT and SoundStorm. Furthermore, we conduct an experiment where the loss is calculated only on a single codebook, which also results in significant performance degradation, confirming the advantages of the proposed dense loss computation.

\begin{table}[htbp]
\centering
\caption{Impact of LLM initialization on WER across datasets. 
Abbreviations: Libri = Librispeech-PC test-clean, 
Seed-en = Seed-TTS test-en, Seed-zh = Seed-TTS test-zh.}
\label{tab:ablation_llm}
\begin{tabular}{lcccc}
\toprule
\multirow{2}{*}{\textbf{Model}} & \multirow{2}{*}{\textbf{Learning Rate}} & \multicolumn{3}{c}{\textbf{WER $\downarrow$}} \\
\cmidrule(lr){3-5}
& & \textbf{Libri} & \textbf{Seed-en} & \textbf{Seed-zh} \\
\midrule
LLM initialization & 1e-4 & \textbf{1.57} & \textbf{1.72} & \textbf{0.89} \\
\midrule
\multirow{4}{*}{Random initialization} & 1e-4  & 2.79 & 2.34 & 1.11 \\
& 2e-4  & 2.52 & 2.43 & 1.01 \\
& 5e-4  & 2.56 & 2.07 & 1.01 \\
& 1e-3  & 2.72 & 2.29 & 1.02 \\
\bottomrule
\end{tabular}
\end{table}

We illustrate the importance of LLM initialization for high intelligibility (i.e., low WER) in \autoref{tab:ablation_llm}. It can be observed that even after extensive learning rate tuning, the WERs of models without LLM initialization are still consistently higher than those of the model with LLM initialization, highlighting the importance of inheriting linguistic knowledge from pre-trained LLMs.

\begin{table}[htbp]
\centering
\caption{Impact of prompt denoising task.}
\label{tab:ablation_denoise}
\begin{tabular}{lccc}
\toprule
\multirow{2}{*}{\textbf{Model}} & \multicolumn{3}{c}{\textbf{Librispeech-PC test-clean}} \\
\cmidrule(lr){2-4}
& \textbf{SIM-o} $\uparrow$ & \textbf{WER} $\downarrow$ & \textbf{UTMOS} $\uparrow$ \\
\midrule
w/o prompt denoise & \textbf{0.697} & 1.57 & 4.23 \\
w/ prompt denoise & 0.668 & \textbf{1.56} & \textbf{4.32} \\
\bottomrule
\end{tabular}
\end{table}

We illustrate the impact of prompt denoising in \autoref{tab:ablation_denoise}. When prompt denoising is enabled, UTMOS improves from 4.23 to 4.32 because the model generates cleaner speech, while SIM-o decreases slightly (0.697 → 0.668) because the generated speech is more standardized than the noisy prompt, which aligns with our design objective of generating high-quality speech from degraded prompts.

\subsubsection{Inference Speed}

We evaluate the inference speed of OmniVoice on an H20 GPU using PyTorch as the inference framework. Following the identical experimental setup in \cite{zhu2025zipvoice,chen2024f5}, we use a 3-second audio prompt to generate 10-second audio clips, and report the real-time factor (RTF) across different batch sizes and inference steps. As shown in \autoref{tab:rtf}, OmniVoice achieves an RTF of 0.0319 with 16 inference steps and a batch size of 1, outperforming the corresponding results of ZipVoice (0.0557) with the same inference step configuration. Notably, OmniVoice also yields better generation quality in the same 16-step setting (shown in \autoref{sec:steps}). When batch inference is employed, OmniVoice attains an even lower RTF of 0.022. In addition, its inference efficiency can be further improved with acceleration techniques such as TensorRT. These results demonstrate the high inference efficiency of OmniVoice.

\begin{table}[htbp]
\centering
\caption{RTF with different inference steps and batch sizes.}
\label{tab:rtf}
\begin{tabular}{ccccc}
\toprule
\multirow{2}{*}{\textbf{Inference Steps}} & \multicolumn{4}{c}{\textbf{Batch Size}} \\
\cmidrule(lr){2-5}
& 1 & 2 & 4 & 8 \\
\midrule
16  & 0.0319 & 0.0263 & 0.0235 & 0.0224 \\
32  & 0.0598 & 0.0486 & 0.0436 & 0.0414 \\
\bottomrule
\end{tabular}
\end{table}

\section{Conclusions}

In this work, we introduce OmniVoice, a massively multilingual zero-shot TTS model supporting over 600 languages with SOTA performance, addressing the critical limitation of narrow language coverage in existing zero-shot TTS systems. The core of OmniVoice is a novel diffusion language model-style single-stage discrete-token NAR framework, which directly maps text to multi-codebook acoustic tokens to circumvent the inherent limitations of conventional two-stage models. To unlock the full potential of this streamlined architecture, we propose a full-codebook random masking strategy to enhance training efficiency and LLM weight initialization to resolve the intelligibility challenge. Trained on a 581k-hour multilingual dataset curated exclusively from open-source resources, OmniVoice achieves the broadest language coverage of zero-shot TTS models to date, delivers SOTA performance on Chinese, English and diverse multilingual benchmarks, exhibits exceptional generalization to low-resource languages, and is equipped with multi-dimensional controllability and high inference efficiency, marking a significant advance in expanding high-quality TTS capabilities to a broad spectrum of the world’s languages.

\clearpage
\bibliographystyle{unsrt}
\bibliography{main} 

\begin{thebibliography}{10}

\bibitem{chen2025neural}
Sanyuan Chen, Chengyi Wang, Yu~Wu, Ziqiang Zhang, Long Zhou, Shujie Liu, Zhuo Chen, Yanqing Liu, Huaming Wang, Jinyu Li, et~al.
\newblock Neural codec language models are zero-shot text to speech synthesizers.
\newblock {\em IEEE Transactions on Audio, Speech and Language Processing}, 2025.

\bibitem{anastassiou2024seed}
Philip Anastassiou, Jiawei Chen, Jitong Chen, Yuanzhe Chen, Zhuo Chen, Ziyi Chen, Jian Cong, Lelai Deng, Chuang Ding, Lu~Gao, et~al.
\newblock Seed-tts: A family of high-quality versatile speech generation models.
\newblock {\em arXiv preprint arXiv:2406.02430}, 2024.

\bibitem{du2024cosyvoice2}
Zhihao Du, Yuxuan Wang, Qian Chen, Xian Shi, Xiang Lv, Tianyu Zhao, Zhifu Gao, Yexin Yang, Changfeng Gao, Hui Wang, et~al.
\newblock Cosyvoice 2: Scalable streaming speech synthesis with large language models.
\newblock {\em arXiv preprint arXiv:2412.10117}, 2024.

\bibitem{guo2024fireredtts}
Hao-Han Guo, Yao Hu, Kun Liu, Fei-Yu Shen, Xu~Tang, Yi-Chen Wu, Feng-Long Xie, Kun Xie, and Kai-Tuo Xu.
\newblock Fireredtts: A foundation text-to-speech framework for industry-level generative speech applications.
\newblock {\em arXiv preprint arXiv:2409.03283}, 2024.

\bibitem{zhou2025indextts2}
Siyi Zhou, Yiquan Zhou, Yi~He, Xun Zhou, Jinchao Wang, Wei Deng, and Jingchen Shu.
\newblock Indextts2: A breakthrough in emotionally expressive and duration-controlled auto-regressive zero-shot text-to-speech.
\newblock {\em arXiv preprint arXiv:2506.21619}, 2025.

\bibitem{jia2025ditar}
Dongya Jia, Zhuo Chen, Jiawei Chen, Chenpeng Du, Jian Wu, Jian Cong, Xiaobin Zhuang, Chumin Li, Zhen Wei, Yuping Wang, et~al.
\newblock Ditar: Diffusion transformer autoregressive modeling for speech generation.
\newblock In {\em International Conference on Machine Learning}, pages 27255--27270. PMLR, 2025.

\bibitem{wang2025spark}
Xinsheng Wang, Mingqi Jiang, Ziyang Ma, Ziyu Zhang, Songxiang Liu, Linqin Li, Zheng Liang, Qixi Zheng, Rui Wang, Xiaoqin Feng, et~al.
\newblock Spark-tts: An efficient llm-based text-to-speech model with single-stream decoupled speech tokens.
\newblock {\em arXiv preprint arXiv:2503.01710}, 2025.

\bibitem{ye2025llasa}
Zhen Ye, Xinfa Zhu, Chi-Min Chan, Xinsheng Wang, Xu~Tan, Jiahe Lei, Yi~Peng, Haohe Liu, Yizhu Jin, Zheqi Dai, et~al.
\newblock Llasa: Scaling train-time and inference-time compute for llama-based speech synthesis.
\newblock {\em arXiv preprint arXiv:2502.04128}, 2025.

\bibitem{song2025distar}
Yakun Song, Xiaobin Zhuang, Jiawei Chen, Zhikang Niu, Guanrou Yang, Chenpeng Du, Dongya Jia, Zhuo Chen, Yuping Wang, Yuxuan Wang, et~al.
\newblock Distar: Diffusion over a scalable token autoregressive representation for speech generation.
\newblock {\em arXiv preprint arXiv:2510.12210}, 2025.

\bibitem{zhou2025voxcpm}
Yixuan Zhou, Guoyang Zeng, Xin Liu, Xiang Li, Renjie Yu, Ziyang Wang, Runchuan Ye, Weiyue Sun, Jiancheng Gui, Kehan Li, et~al.
\newblock Voxcpm: Tokenizer-free tts for context-aware speech generation and true-to-life voice cloning.
\newblock {\em arXiv preprint arXiv:2509.24650}, 2025.

\bibitem{cui2025glm}
Jiayan Cui, Zhihan Yang, Naihan Li, Jiankun Tian, Xingyu Ma, Yi~Zhang, Guangyu Chen, Runxuan Yang, Yuqing Cheng, Yizhi Zhou, et~al.
\newblock Glm-tts technical report.
\newblock {\em arXiv preprint arXiv:2512.14291}, 2025.

\bibitem{le2023voicebox}
Matthew Le, Apoorv Vyas, Bowen Shi, Brian Karrer, Leda Sari, Rashel Moritz, Mary Williamson, Vimal Manohar, Yossi Adi, Jay Mahadeokar, et~al.
\newblock Voicebox: Text-guided multilingual universal speech generation at scale.
\newblock {\em Advances in neural information processing systems}, 36:14005--14034, 2023.

\bibitem{eskimez2024e2}
Sefik~Emre Eskimez, Xiaofei Wang, Manthan Thakker, Canrun Li, Chung-Hsien Tsai, Zhen Xiao, Hemin Yang, Zirun Zhu, Min Tang, Xu~Tan, et~al.
\newblock E2 tts: Embarrassingly easy fully non-autoregressive zero-shot tts.
\newblock In {\em 2024 IEEE Spoken Language Technology Workshop (SLT)}, pages 682--689. IEEE, 2024.

\bibitem{chen2024f5}
Yushen Chen, Zhikang Niu, Ziyang Ma, Keqi Deng, Chunhui Wang, Jian Zhao, Kai Yu, and Xie Chen.
\newblock F5-tts: A fairytaler that fakes fluent and faithful speech with flow matching.
\newblock {\em arXiv preprint arXiv:2410.06885}, 2024.

\bibitem{yang2025pseudo}
Yifan Yang, Shujie Liu, Jinyu Li, Yuxuan Hu, Haibin Wu, Hui Wang, Jianwei Yu, Lingwei Meng, Haiyang Sun, Yanqing Liu, et~al.
\newblock Pseudo-autoregressive neural codec language models for efficient zero-shot text-to-speech synthesis.
\newblock In {\em Proceedings of the 33rd ACM International Conference on Multimedia}, pages 9316--9325, 2025.

\bibitem{zhu2025zipvoice}
Han Zhu, Wei Kang, Zengwei Yao, Liyong Guo, Fangjun Kuang, Zhaoqing Li, Weiji Zhuang, Long Lin, and Daniel Povey.
\newblock Zipvoice: Fast and high-quality zero-shot text-to-speech with flow matching.
\newblock {\em arXiv preprint arXiv:2506.13053}, 2025.

\bibitem{zhu2025zipvoicedialog}
Han Zhu, Wei Kang, Liyong Guo, Zengwei Yao, Fangjun Kuang, Weiji Zhuang, Zhaoqing Li, Zhifeng Han, Dong Zhang, Xin Zhang, et~al.
\newblock Zipvoice-dialog: Non-autoregressive spoken dialogue generation with flow matching.
\newblock {\em arXiv preprint arXiv:2507.09318}, 2025.

\bibitem{jiang2025megatts}
Ziyue Jiang, Yi~Ren, Ruiqi Li, Shengpeng Ji, Boyang Zhang, Zhenhui Ye, Chen Zhang, Bai Jionghao, Xiaoda Yang, Jialong Zuo, et~al.
\newblock Megatts 3: Sparse alignment enhanced latent diffusion transformer for zero-shot speech synthesis.
\newblock {\em arXiv preprint arXiv:2502.18924}, 2025.

\bibitem{wang2025maskgct}
Yuancheng Wang, Haoyue Zhan, Liwei Liu, Ruihong Zeng, Haotian Guo, Jiachen Zheng, Qiang Zhang, Xueyao Zhang, Shunsi Zhang, and Zhizheng Wu.
\newblock Mask{GCT}: Zero-shot text-to-speech with masked generative codec transformer.
\newblock In {\em The Thirteenth International Conference on Learning Representations}, 2025.

\bibitem{yang2025measuring}
Yifan Yang, Bing Han, Hui Wang, Long Zhou, Wei Wang, Mingyu Cui, Xu~Tan, and Xie Chen.
\newblock Measuring prosody diversity in zero-shot tts: A new metric, benchmark, and exploration.
\newblock {\em arXiv preprint arXiv:2509.19928}, 2025.

\bibitem{gallego2025single}
Gerard~I G{\'a}llego, Roy Fejgin, Chunghsin Yeh, Xiaoyu Liu, and Gautam Bhattacharya.
\newblock Single-stage tts with masked audio token modeling and semantic knowledge distillation.
\newblock In {\em ICASSP 2025-2025 IEEE International Conference on Acoustics, Speech and Signal Processing (ICASSP)}, pages 1--5. IEEE, 2025.

\bibitem{sahoo2024simple}
Subham Sahoo, Marianne Arriola, Yair Schiff, Aaron Gokaslan, Edgar Marroquin, Justin Chiu, Alexander Rush, and Volodymyr Kuleshov.
\newblock Simple and effective masked diffusion language models.
\newblock {\em Advances in Neural Information Processing Systems}, 37:130136--130184, 2024.

\bibitem{vaswani2017attention}
Ashish Vaswani, Noam Shazeer, Niki Parmar, Jakob Uszkoreit, Llion Jones, Aidan~N Gomez, {\L}ukasz Kaiser, and Illia Polosukhin.
\newblock Attention is all you need.
\newblock {\em Advances in neural information processing systems}, 30, 2017.

\bibitem{nie2025large}
Shen Nie, Fengqi Zhu, Zebin You, Xiaolu Zhang, Jingyang Ou, Jun Hu, JUN ZHOU, Yankai Lin, Ji-Rong Wen, and Chongxuan Li.
\newblock Large language diffusion models.
\newblock In {\em The Thirty-ninth Annual Conference on Neural Information Processing Systems}, 2025.

\bibitem{ye2025dream}
Jiacheng Ye, Zhihui Xie, Lin Zheng, Jiahui Gao, Zirui Wu, Xin Jiang, Zhenguo Li, and Lingpeng Kong.
\newblock Dream 7b: Diffusion large language models.
\newblock {\em arXiv preprint arXiv:2508.15487}, 2025.

\bibitem{borsos2023soundstorm}
Zal{\'a}n Borsos, Matt Sharifi, Damien Vincent, Eugene Kharitonov, Neil Zeghidour, and Marco Tagliasacchi.
\newblock Soundstorm: Efficient parallel audio generation.
\newblock {\em arXiv preprint arXiv:2305.09636}, 2023.

\bibitem{zhang2025advanced}
Leying Zhang, Wangyou Zhang, Zhengyang Chen, and Yanmin Qian.
\newblock Advanced zero-shot text-to-speech for background removal and preservation with controllable masked speech prediction.
\newblock In {\em ICASSP 2025-2025 IEEE International Conference on Acoustics, Speech and Signal Processing (ICASSP)}, pages 1--5. IEEE, 2025.

\bibitem{wang2024investigation}
Xiaofei Wang, Sefik~Emre Eskimez, Manthan Thakker, Hemin Yang, Zirun Zhu, Min Tang, Yufei Xia, Jinzhu Li, Sheng Zhao, Jinyu Li, et~al.
\newblock An investigation of noise robustness for flow-matching-based zero-shot tts.
\newblock {\em arXiv preprint arXiv:2406.05699}, 2024.

\bibitem{hu2026voicesculptor}
Jingbin Hu, Huakang Chen, Linhan Ma, Dake Guo, Qirui Zhan, Wenhao Li, Haoyu Zhang, Kangxiang Xia, Ziyu Zhang, Wenjie Tian, et~al.
\newblock Voicesculptor: Your voice, designed by you.
\newblock {\em arXiv preprint arXiv:2601.10629}, 2026.

\bibitem{liao2025nvspeech}
Huan Liao, Qinke Ni, Yuancheng Wang, Yiheng Lu, Haoyue Zhan, Pengyuan Xie, Qiang Zhang, and Zhizheng Wu.
\newblock Nvspeech: An integrated and scalable pipeline for human-like speech modeling with paralinguistic vocalizations.
\newblock {\em arXiv preprint arXiv:2508.04195}, 2025.

\bibitem{deng2025indextts}
Wei Deng, Siyi Zhou, Jingchen Shu, Jinchao Wang, and Lu~Wang.
\newblock Indextts: An industrial-level controllable and efficient zero-shot text-to-speech system.
\newblock {\em arXiv preprint arXiv:2502.05512}, 2025.

\bibitem{yang2025qwen3}
An~Yang, Anfeng Li, Baosong Yang, Beichen Zhang, Binyuan Hui, Bo~Zheng, Bowen Yu, Chang Gao, Chengen Huang, Chenxu Lv, et~al.
\newblock Qwen3 technical report.
\newblock {\em arXiv preprint arXiv:2505.09388}, 2025.

\bibitem{du2025cosyvoice}
Zhihao Du, Changfeng Gao, Yuxuan Wang, Fan Yu, Tianyu Zhao, Hao Wang, Xiang Lv, Hui Wang, Chongjia Ni, Xian Shi, et~al.
\newblock Cosyvoice 3: Towards in-the-wild speech generation via scaling-up and post-training.
\newblock {\em arXiv preprint arXiv:2505.17589}, 2025.

\bibitem{casanova2022yourtts}
Edresson Casanova, Julian Weber, Christopher~D Shulby, Arnaldo~Candido Junior, Eren G{\"o}lge, and Moacir~A Ponti.
\newblock Yourtts: Towards zero-shot multi-speaker tts and zero-shot voice conversion for everyone.
\newblock In {\em International conference on machine learning}, pages 2709--2720. PMLR, 2022.

\bibitem{casanova2024xtts}
Edresson Casanova, Kelly Davis, Eren G{\"o}lge, G{\"o}rkem G{\"o}knar, Iulian Gulea, Logan Hart, Aya Aljafari, Joshua Meyer, Reuben Morais, Samuel Olayemi, et~al.
\newblock Xtts: a massively multilingual zero-shot text-to-speech model.
\newblock In {\em Proc. Interspeech 2024}, pages 4978--4982, 2024.

\bibitem{zheng2025voicecraft}
Zhisheng Zheng, Puyuan Peng, Anuj Diwan, Cong~Phuoc Huynh, Xiaohang Sun, Zhu Liu, Vimal Bhat, and David Harwath.
\newblock Voicecraft-x: Unifying multilingual, voice-cloning speech synthesis and speech editing.
\newblock In {\em Proceedings of the 2025 Conference on Empirical Methods in Natural Language Processing}, pages 2737--2756, 2025.

\bibitem{chen2026habibi}
Yushen Chen, Junzhe Liu, Yujie Tu, Zhikang Niu, Yuzhe Liang, Kai Yu, Chunyu Qiang, Chen Zhang, and Xie Chen.
\newblock Habibi: Laying the open-source foundation of unified-dialectal arabic speech synthesis.
\newblock {\em arXiv preprint arXiv:2601.13802}, 2026.

\bibitem{zhao2026lemas}
Zhiyuan Zhao, Lijian Lin, Ye~Zhu, Kai Xie, Yunfei Liu, and Yu~Li.
\newblock Lemas: Large a 150k-hour large-scale extensible multilingual audio suite with generative speech models.
\newblock {\em arXiv preprint arXiv:2601.04233}, 2026.

\bibitem{pratap2024scaling}
Vineel Pratap, Andros Tjandra, Bowen Shi, Paden Tomasello, Arun Babu, Sayani Kundu, Ali Elkahky, Zhaoheng Ni, Apoorv Vyas, Maryam Fazel-Zarandi, et~al.
\newblock Scaling speech technology to 1,000+ languages.
\newblock {\em Journal of Machine Learning Research}, 25(97):1--52, 2024.

\bibitem{chatterboxtts2025}
{Resemble AI}.
\newblock {Chatterbox-TTS}.
\newblock \url{https://github.com/resemble-ai/chatterbox}, 2025.
\newblock GitHub repository.

\bibitem{liao2024fish}
Shijia Liao, Yuxuan Wang, Tianyu Li, Yifan Cheng, Ruoyi Zhang, Rongzhi Zhou, and Yijin Xing.
\newblock Fish-speech: Leveraging large language models for advanced multilingual text-to-speech synthesis.
\newblock {\em arXiv preprint arXiv:2411.01156}, 2024.

\bibitem{hu2026qwen3}
Hangrui Hu, Xinfa Zhu, Ting He, Dake Guo, Bin Zhang, Xiong Wang, Zhifang Guo, Ziyue Jiang, Hongkun Hao, Zishan Guo, et~al.
\newblock Qwen3-tts technical report.
\newblock {\em arXiv preprint arXiv:2601.15621}, 2026.

\bibitem{li2026indextts}
Yunpei Li, Xun Zhou, Jinchao Wang, Lu~Wang, Yong Wu, Siyi Zhou, Yiquan Zhou, and Jingchen Shu.
\newblock Indextts 2.5 technical report.
\newblock {\em arXiv preprint arXiv:2601.03888}, 2026.

\bibitem{nakata2025sidon}
Wataru Nakata, Yuki Saito, Yota Ueda, and Hiroshi Saruwatari.
\newblock Sidon: Fast and robust open-source multilingual speech restoration for large-scale dataset cleansing.
\newblock {\em arXiv preprint arXiv:2509.17052}, 2025.

\bibitem{ye2025scalable}
Runchuan Ye, Yixuan Zhou, Renjie Yu, Zijian Lin, Kehan Li, Xiang Li, Xin Liu, Guoyang Zeng, and Zhiyong Wu.
\newblock A scalable pipeline for enabling non-verbal speech generation and understanding.
\newblock {\em arXiv preprint arXiv:2508.05385}, 2025.

\bibitem{he2025emilia}
Haorui He, Zengqiang Shang, Chaoren Wang, Xuyuan Li, Yicheng Gu, Hua Hua, Liwei Liu, Chen Yang, Jiaqi Li, Peiyang Shi, et~al.
\newblock Emilia: A large-scale, extensive, multilingual, and diverse dataset for speech generation.
\newblock {\em IEEE Transactions on Audio, Speech and Language Processing}, 2025.

\bibitem{higgsaudio2025}
{Boson AI}.
\newblock {Higgs Audio V2: Redefining Expressiveness in Audio Generation}.
\newblock \url{https://github.com/boson-ai/higgs-audio}, 2025.
\newblock GitHub repository. Release blog available at \url{https://www.boson.ai/blog/higgs-audio-v2}.

\bibitem{loshchilov2017decoupled}
Ilya Loshchilov and Frank Hutter.
\newblock Decoupled weight decay regularization.
\newblock {\em arXiv preprint arXiv:1711.05101}, 2017.

\bibitem{ho2021classifierfree}
Jonathan Ho and Tim Salimans.
\newblock Classifier-free diffusion guidance.
\newblock In {\em NeurIPS 2021 Workshop on Deep Generative Models and Downstream Applications}, 2021.

\bibitem{meister2023librispeech}
Aleksandr Meister, Matvei Novikov, Nikolay Karpov, Evelina Bakhturina, Vitaly Lavrukhin, and Boris Ginsburg.
\newblock Librispeech-pc: Benchmark for evaluation of punctuation and capitalization capabilities of end-to-end asr models.
\newblock In {\em 2023 IEEE automatic speech recognition and understanding workshop (ASRU)}, pages 1--7. IEEE, 2023.

\bibitem{zhang2025minimax}
Bowen Zhang, Congchao Guo, Geng Yang, Hang Yu, Haozhe Zhang, Heidi Lei, Jialong Mai, Junjie Yan, Kaiyue Yang, Mingqi Yang, et~al.
\newblock Minimax-speech: Intrinsic zero-shot text-to-speech with a learnable speaker encoder.
\newblock {\em arXiv preprint arXiv:2505.07916}, 2025.

\bibitem{conneau2023fleurs}
Alexis Conneau, Min Ma, Simran Khanuja, Yu~Zhang, Vera Axelrod, Siddharth Dalmia, Jason Riesa, Clara Rivera, and Ankur Bapna.
\newblock Fleurs: Few-shot learning evaluation of universal representations of speech.
\newblock In {\em 2022 IEEE Spoken Language Technology Workshop (SLT)}, pages 798--805. IEEE, 2023.

\bibitem{chen2022wavlm}
Sanyuan Chen, Chengyi Wang, Zhengyang Chen, Yu~Wu, Shujie Liu, Zhuo Chen, Jinyu Li, Naoyuki Kanda, Takuya Yoshioka, Xiong Xiao, et~al.
\newblock Wavlm: Large-scale self-supervised pre-training for full stack speech processing.
\newblock {\em IEEE Journal of Selected Topics in Signal Processing}, 16(6):1505--1518, 2022.

\bibitem{desplanques2020ecapa}
Brecht Desplanques, Jenthe Thienpondt, and Kris Demuynck.
\newblock Ecapa-tdnn: Emphasized channel attention, propagation and aggregation in tdnn based speaker verification.
\newblock In {\em Proc. Interspeech 2020}, pages 3830--3834, 2020.

\bibitem{hsu2021hubert}
Wei-Ning Hsu, Benjamin Bolte, Yao-Hung~Hubert Tsai, Kushal Lakhotia, Ruslan Salakhutdinov, and Abdelrahman Mohamed.
\newblock Hubert: Self-supervised speech representation learning by masked prediction of hidden units.
\newblock {\em IEEE/ACM transactions on audio, speech, and language processing}, 29:3451--3460, 2021.

\bibitem{gao2022paraformer}
Zhifu Gao, ShiLiang Zhang, Ian McLoughlin, and Zhijie Yan.
\newblock Paraformer: Fast and accurate parallel transformer for non-autoregressive end-to-end speech recognition.
\newblock In {\em Proc. Interspeech 2022}, pages 2063--2067, 2022.

\bibitem{omnilingual2025omnilingual}
ASR Omnilingual, Gil Keren, Artyom Kozhevnikov, Yen Meng, Christophe Ropers, Matthew Setzler, Skyler Wang, Ife Adebara, Michael Auli, Can Balioglu, et~al.
\newblock Omnilingual asr: Open-source multilingual speech recognition for 1600+ languages.
\newblock {\em arXiv preprint arXiv:2511.09690}, 2025.

\bibitem{radford2023robust}
Alec Radford, Jong~Wook Kim, Tao Xu, Greg Brockman, Christine McLeavey, and Ilya Sutskever.
\newblock Robust speech recognition via large-scale weak supervision.
\newblock In {\em International conference on machine learning}, pages 28492--28518. PMLR, 2023.

\bibitem{saeki2022utmos}
Takaaki Saeki, Detai Xin, Wataru Nakata, Tomoki Koriyama, Shinnosuke Takamichi, and Hiroshi Saruwatari.
\newblock Utmos: Utokyo-sarulab system for voicemos challenge 2022.
\newblock {\em Interspeech 2022}, 2022.

\bibitem{an2024funaudiollm}
Keyu An, Qian Chen, Chong Deng, Zhihao Du, Changfeng Gao, Zhifu Gao, Yue Gu, Ting He, Hangrui Hu, Kai Hu, et~al.
\newblock Funaudiollm: Voice understanding and generation foundation models for natural interaction between humans and llms.
\newblock {\em arXiv preprint arXiv:2407.04051}, 2024.

\bibitem{zen2019libritts}
Heiga Zen, Viet Dang, Rob Clark, Yu~Zhang, Ron~J Weiss, Ye~Jia, Zhifeng Chen, and Yonghui Wu.
\newblock Libritts: A corpus derived from librispeech for text-to-speech.
\newblock In {\em Proc. Interspeech 2019}, pages 1526--1530, 2019.

\bibitem{ardila2020common}
Rosana Ardila, Megan Branson, Kelly Davis, Michael Kohler, Josh Meyer, Michael Henretty, Reuben Morais, Lindsay Saunders, Francis Tyers, and Gregor Weber.
\newblock Common voice: A massively-multilingual speech corpus.
\newblock In {\em Proceedings of the Twelfth Language Resources and Evaluation Conference}, pages 4218--4222, 2020.

\bibitem{yang2025gigaspeech}
Yifan Yang, Zheshu Song, Jianheng Zhuo, Mingyu Cui, Jinpeng Li, Bo~Yang, Yexing Du, Ziyang Ma, Xunying Liu, Ziyuan Wang, et~al.
\newblock Gigaspeech 2: An evolving, large-scale and multi-domain asr corpus for low-resource languages with automated crawling, transcription and refinement.
\newblock In {\em Proceedings of the 63rd Annual Meeting of the Association for Computational Linguistics (Volume 1: Long Papers)}, pages 2673--2686, 2025.

\bibitem{koluguri2025granary}
Nithin~Rao Koluguri, Monica Sekoyan, George Zelenfroynd, Sasha Meister, Shuoyang Ding, Sofia Kostandian, He~Huang, Nikolay Karpov, Jagadeesh Balam, Vitaly Lavrukhin, et~al.
\newblock Granary: Speech recognition and translation dataset in 25 european languages.
\newblock {\em arXiv preprint arXiv:2505.13404}, 2025.

\bibitem{pfisterer2025eurospeech}
Samuel Pfisterer, Florian Gr{\"o}tschla, Luca~A Lanzend{\"o}rfer, Florian Yan, and Roger Wattenhofer.
\newblock Eurospeech: A multilingual speech corpus.
\newblock {\em arXiv preprint arXiv:2510.00514}, 2025.

\bibitem{sankar2024indicvoices}
Ashwin Sankar, Srija Anand, Praveen Varadhan, Sherry Thomas, Mehak Singal, Shridhar Kumar, Deovrat Mehendale, Aditi Krishana, Giri Raju, and Mitesh Khapra.
\newblock Indicvoices-r: Unlocking a massive multilingual multi-speaker speech corpus for scaling indian tts.
\newblock {\em Advances in Neural Information Processing Systems}, 37:68161--68182, 2024.

\bibitem{kumar2023towards}
Gokul~Karthik Kumar, SV~Praveen, Pratyush Kumar, Mitesh~M Khapra, and Karthik Nandakumar.
\newblock Towards building text-to-speech systems for the next billion users.
\newblock In {\em Icassp 2023-2023 ieee international conference on acoustics, speech and signal processing (icassp)}, pages 1--5. IEEE, 2023.

\bibitem{srinivasa2024rasa}
Praveen Srinivasa~Varadhan, Ashwin Sankar, Giri Raju, and Mitesh~M Khapra.
\newblock Rasa: Building expressive speech synthesis systems for indian languages in low-resource settings.
\newblock In {\em Proc. Interspeech 2024}, pages 1830--1834, 2024.

\bibitem{vu2025zeroshottexttospeechvietnamese}
Thi Vu, Linh~The Nguyen, and Dat~Quoc Nguyen.
\newblock Zero-shot text-to-speech for vietnamese.
\newblock In {\em Proceedings of ACL}, 2025.

\bibitem{oliveira2023cml}
Frederico~S Oliveira, Edresson Casanova, Arnaldo~Candido Junior, Anderson~S Soares, and Arlindo~R Galv{\~a}o~Filho.
\newblock Cml-tts: A multilingual dataset for speech synthesis in low-resource languages.
\newblock In {\em International Conference on Text, Speech, and Dialogue}, pages 188--199. Springer, 2023.

\bibitem{li2025wenetspeech}
Longhao Li, Zhao Guo, Hongjie Chen, Yuhang Dai, Ziyu Zhang, Hongfei Xue, Tianlun Zuo, Chengyou Wang, Shuiyuan Wang, Jie Li, et~al.
\newblock Wenetspeech-yue: A large-scale cantonese speech corpus with multi-dimensional annotation.
\newblock {\em arXiv preprint arXiv:2509.03959}, 2025.

\bibitem{dai2025wenetspeech}
Yuhang Dai, Ziyu Zhang, Shuai Wang, Longhao Li, Zhao Guo, Tianlun Zuo, Shuiyuan Wang, Hongfei Xue, Chengyou Wang, Qing Wang, et~al.
\newblock Wenetspeech-chuan: A large-scale sichuanese corpus with rich annotation for dialectal speech processing.
\newblock {\em arXiv preprint arXiv:2509.18004}, 2025.

\bibitem{tang2021kespeech}
Zhiyuan Tang, Dong Wang, Yanguang Xu, Jianwei Sun, Xiaoning Lei, Shuaijiang Zhao, Cheng Wen, Xingjun Tan, Chuandong Xie, Shuran Zhou, et~al.
\newblock Kespeech: An open source speech dataset of mandarin and its eight subdialects.
\newblock In {\em Thirty-fifth conference on neural information processing systems datasets and benchmarks track (Round 2)}, 2021.

\bibitem{shi2021accented}
Xian Shi, Fan Yu, Yizhou Lu, Yuhao Liang, Qiangze Feng, Daliang Wang, Yanmin Qian, and Lei Xie.
\newblock The accented english speech recognition challenge 2020: open datasets, tracks, baselines, results and methods.
\newblock In {\em ICASSP 2021-2021 IEEE International Conference on Acoustics, Speech and Signal Processing (ICASSP)}, pages 6918--6922. IEEE, 2021.

\bibitem{bang2020ksponspeech}
Jeong-Uk Bang, Seung Yun, Seung-Hi Kim, Mu-Yeol Choi, Min-Kyu Lee, Yeo-Jeong Kim, Dong-Hyun Kim, Jun Park, Young-Jik Lee, and Sang-Hun Kim.
\newblock Ksponspeech: Korean spontaneous speech corpus for automatic speech recognition.
\newblock {\em Applied Sciences}, 10(19):6936, 2020.

\bibitem{fujimoto2016reazonspeech}
YYDMS Fujimoto.
\newblock Reazonspeech: A free and massive corpus for japanese asr, 2016.

\bibitem{li2025aishell6}
Cancan Li, Fei Su, Juan Liu, Hui Bu, Yulong Wan, Hongbin Suo, and Ming Li.
\newblock Aishell6-whisper: A chinese mandarin audio-visual whisper speech dataset with speech recognition baselines.
\newblock {\em arXiv preprint arXiv:2509.23833}, 2025.

\bibitem{diwan2025scaling}
Anuj Diwan, Zhisheng Zheng, David Harwath, and Eunsol Choi.
\newblock Scaling rich style-prompted text-to-speech datasets.
\newblock In {\em Proceedings of the 2025 Conference on Empirical Methods in Natural Language Processing}, pages 3639--3659, 2025.

\bibitem{chen2025seniortalkchineseconversationdataset}
Yang Chen, Hui Wang, Shiyao Wang, Junyang Chen, Jiabei He, Jiaming Zhou, Xi~Yang, Yequan Wang, Yonghua Lin, and Yong Qin.
\newblock Seniortalk: A chinese conversation dataset with rich annotations for super-aged seniors, 2025.

\bibitem{zhou2024childmandarin}
Jiaming Zhou, Shiyao Wang, Shiwan Zhao, Jiabei He, Haoqin Sun, Hui Wang, Cheng Liu, Aobo Kong, Yujie Guo, and Yong Qin.
\newblock Childmandarin: A comprehensive mandarin speech dataset for young children aged 3-5.
\newblock {\em arXiv preprint arXiv:2409.18584}, 2024.

\bibitem{mussakhojayeva22_interspeech}
Saida Mussakhojayeva, Yerbolat Khassanov, and Huseyin {Atakan Varol}.
\newblock Ksc2: An industrial-scale open-source kazakh speech corpus.
\newblock In {\em Interspeech 2022}, pages 1367--1371, 2022.

\bibitem{kulkarni2023clartts}
Ajinkya Kulkarni, Atharva Kulkarni, Sara Abedalmon'em~Mohammad Shatnawi, and Hanan Aldarmaki.
\newblock Clartts: An open-source classical arabic text-to-speech corpus.
\newblock In {\em Proc. Interspeech 2023}, pages 5511--5515, 2023.

\bibitem{toyin2025arvoice}
Hawau Toyin, Rufael Marew, Humaid Alblooshi, Samar~M Magdy, and Hanan Aldarmaki.
\newblock Arvoice: A multi-speaker dataset for arabic speech synthesis.
\newblock In {\em Proc. Interspeech 2025}, pages 4808--4812, 2025.

\bibitem{mussakhojayeva2022kazakhtts2}
Saida Mussakhojayeva, Yerbolat Khassanov, and Huseyin~Atakan Varol.
\newblock Kazakhtts2: Extending the open-source kazakh tts corpus with more data, speakers, and topics.
\newblock In {\em Proceedings of the Thirteenth Language Resources and Evaluation Conference}, pages 5404--5411, 2022.

\bibitem{liu2022imut}
Zhiqiang Liu, Zhiqiang Ma, Xiaoxu Zhang, Caijilahu Bao, Xiulan Xie, and Fangyuan Zhu.
\newblock Imut-mc: A speech corpus for mongolian speech recognition.
\newblock {\em China Sci. Data}, 7:13, 2022.

\bibitem{ali2016mgb}
Ahmed Ali, Peter Bell, James Glass, Yacine Messaoui, Hamdy Mubarak, Steve Renals, and Yifan Zhang.
\newblock The mgb-2 challenge: Arabic multi-dialect broadcast media recognition.
\newblock In {\em 2016 IEEE Spoken Language Technology Workshop (SLT)}, pages 279--284. IEEE, 2016.

\bibitem{ali2017speech}
Ahmed Ali, Stephan Vogel, and Steve Renals.
\newblock Speech recognition challenge in the wild: Arabic mgb-3.
\newblock In {\em 2017 IEEE Automatic Speech Recognition and Understanding Workshop (ASRU)}, pages 316--322. IEEE, 2017.

\bibitem{ali2019mgb}
Ahmed Ali, Suwon Shon, Younes Samih, Hamdy Mubarak, Ahmed Abdelali, James Glass, Steve Renals, and Khalid Choukri.
\newblock The mgb-5 challenge: Recognition and dialect identification of dialectal arabic speech.
\newblock In {\em 2019 IEEE Automatic Speech Recognition and Understanding Workshop (ASRU)}, pages 1026--1033. IEEE, 2019.

\bibitem{alharbi2024sada}
Sadeen Alharbi, Areeb Alowisheq, Zolt{\'a}n T{\"u}ske, Kareem Darwish, Abdullah Alrajeh, Abdulmajeed Alrowithi, Aljawharah~Bin Tamran, Asma Ibrahim, Raghad Aloraini, Raneem Alnajim, et~al.
\newblock Sada: Saudi audio dataset for arabic.
\newblock In {\em ICASSP 2024-2024 IEEE International Conference on Acoustics, Speech and Signal Processing (ICASSP)}, pages 10286--10290. IEEE, 2024.

\bibitem{al2023masc}
Mohammad Al-Fetyani, Muhammad Al-Barham, Gheith Abandah, Adham Alsharkawi, and Maha Dawas.
\newblock Masc: Massive arabic speech corpus.
\newblock In {\em 2022 IEEE Spoken Language Technology Workshop (SLT)}, pages 1006--1013. IEEE, 2023.

\bibitem{al2024mixat}
Maryam~Khalifa Al~Ali and Hanan Aldarmaki.
\newblock Mixat: A data set of bilingual emirati-english speech.
\newblock In {\em Proceedings of the 3rd Annual Meeting of the Special Interest Group on Under-resourced Languages@ LREC-COLING 2024}, pages 222--226, 2024.

\bibitem{nict-tib1}
Kak Soky, Zhuo Gong, and Sheng Li.
\newblock {NICT-Tib1: A Public Speech Corpus of Lhasa Dialect for Benchmarking Tibetan Language Speech Recognition Systems}.
\newblock In {\em Proc. O-COCOSDA}, pages 1--5, 2022.

\bibitem{TIBMDMUC}
Yue Zhao, Xiaona Xu, Jianjian Yue, Wei Song, Xiali Li, Licheng Wu, and Qiang Ji.
\newblock An open speech resource for tibetan multi-dialect and multitask recognition.
\newblock {\em International Journal of Computational Science and Engineering}, 22(2/3):297--304, 2020.

\bibitem{li2022xbmu}
Senyan Li, Guanyu Li, and Jiewen Ning.
\newblock Xbmu-amdo31: An open source of amdo tibetan speech database and speech recognition baseline system.
\newblock In {\em National Conference on Man-Machine Speech Communication, NCMMSC2022}, 2022.

\bibitem{Tibetan-Greetings}
Linfei Lu, Jiaxin Pang, Stansencuo, Buwonglam, and Linting Huang.
\newblock Tibetan greetings.
\newblock \url{http://www.openslr.org/149/}.

\end{thebibliography}


\appendix

\section{Complete List of Multilingual Training Data}
\label{sec:data_list}

We used the following datasets to train OmniVoice:

Emilia~\cite{he2025emilia}, Emilia-YODAS~\cite{he2025emilia}, LibriTTS~\cite{zen2019libritts}, Common Voice~\cite{ardila2020common}, VoxBox~\cite{wang2025spark}, Meta Omnilingual ASR Corpus~\cite{omnilingual2025omnilingual}, FLEURS~\cite{conneau2023fleurs}, GigaSpeech 2~\cite{yang2025gigaspeech}, YODAS-Granary~\cite{koluguri2025granary}, EuroSpeech~\cite{pfisterer2025eurospeech}, IndicVoices-R~\cite{sankar2024indicvoices}, IndicTTS~\cite{kumar2023towards}, Rasa~\cite{srinivasa2024rasa}, PhoAudiobook~\cite{vu2025zeroshottexttospeechvietnamese}, viVoice\footnote{\url{https://huggingface.co/datasets/capleaf/viVoice}}, CML-TTS~\cite{oliveira2023cml}, Wenetspeech-yue~\cite{li2025wenetspeech}, Wenetspeech-chuan~\cite{dai2025wenetspeech}, Kespeech~\cite{tang2021kespeech}, AESRC~\cite{shi2021accented}, Ksponspeech~\cite{bang2020ksponspeech}, Reazonspeech~\cite{fujimoto2016reazonspeech}, AISHELL6-whisper~\cite{li2025aishell6}, ParaSpeechCaps~\cite{diwan2025scaling}, Nvspeech~\cite{liao2025nvspeech}, NonVerbalSpeech-38K~\cite{ye2025scalable}, SeniorTalk~\cite{chen2025seniortalkchineseconversationdataset}, ChildMandarin~\cite{zhou2024childmandarin}, KSC2~\cite{mussakhojayeva22_interspeech}, ClArTTS~\cite{kulkarni2023clartts}, ArVoice~\cite{toyin2025arvoice}, KazakhTTS2~\cite{mussakhojayeva2022kazakhtts2}, RuLS\footnote{\url{https://www.openslr.org/96/}}, IMUT-MC~\cite{liu2022imut}, HUI-Audio-Corpus-German\footnote{\url{https://huggingface.co/datasets/Paradoxia/opendata-iisys-hui}}, MGB-2~\cite{ali2016mgb}, MGB-3~\cite{ali2017speech}, MGB-5~\cite{ali2019mgb}, SADA~\cite{alharbi2024sada}, MASC~\cite{al2023masc}, Mixat~\cite{al2024mixat}, Arabic datasets found from \cite{chen2026habibi} (darija speech to text\footnote{\url{https://huggingface.co/datasets/adiren7/darija_speech_to_text}}, DarijaTTS-clean\footnote{\url{https://huggingface.co/datasets/KandirResearch/DarijaTTS-clean}}, Jordan-Audio\footnote{\url{https://huggingface.co/datasets/nadsoft/Jordan-Audio}}, UAE 100K\footnote{\url{https://huggingface.co/datasets/AhmedBadawy11/UAE_100K}}), NICT-Tib1~\cite{nict-tib1}, TIBMD@MUC~\cite{TIBMDMUC}, XBMU-AMDO3~\cite{li2022xbmu}, Tibetan-Greetings~\cite{Tibetan-Greetings}.

\section{Results of Different Inference Steps}
\label{sec:steps}

In \autoref{tab:steps}, we evaluate the multilingual version OmniVoice model with different inference steps on English and Chinese benchmarks.

\begin{table}[htbp]
\caption{Objective evaluation results of OmniVoice with different inference steps.}
\label{tab:steps}
\centering
\resizebox{\columnwidth}{!}{
\begin{tabular}{cccccccccc}
\toprule
& \multicolumn{3}{c}{\textbf{LibriSpeech-PC test-clean}} & \multicolumn{3}{c}{\textbf{Seed-TTS test-en}} & \multicolumn{3}{c}{\textbf{Seed-TTS test-zh}} \\
\cmidrule(lr){2-4} \cmidrule(lr){5-7} \cmidrule(lr){8-10}
\textbf{Steps} & \textbf{SIM-o $\uparrow$} & \textbf{WER $\downarrow$} & \textbf{UTMOS $\uparrow$} & \textbf{SIM-o $\uparrow$} & \textbf{WER $\downarrow$} & \textbf{UTMOS $\uparrow$} & \textbf{SIM-o $\uparrow$} & \textbf{WER $\downarrow$} & \textbf{UTMOS $\uparrow$} \\
\midrule
64 & 0.729 & 1.28 & 4.30 & 0.742 & 1.60 & 3.92 & 0.777 & 0.81 & 3.13 \\
32 & 0.729 & 1.30 & 4.28 & 0.741 & 1.53 & 3.91 & 0.777 & 0.84 & 3.11 \\
16 & 0.728 & 1.50 & 4.23 & 0.735 & 1.72 & 3.92 & 0.773 & 0.99 & 3.00 \\
8  & 0.713 & 2.02 & 4.02 & 0.716 & 1.94 & 3.59 & 0.756 & 1.58 & 2.72 \\
\bottomrule
\end{tabular}
}
\end{table}

\normalsize

\section{Detailed Results on FLEURS-Multilingual-102}
\label{sec:appendix_fleurs}

Per-language CERs of of ground-truth and OmniVoice on FLEURS-Multilingual-102 is shown in \autoref{fig:fleurs_result}.

\setlength{\LTcapwidth}{\textwidth}
\small
\begin{longtable}{ccc|ccc}
\caption{Per-language CERs of ground-truth and OmniVoice on FLEURS-Multilingual-102}
\label{tab:fleurs_per_lang_cer} \\
\hline
\textbf{Language} & \textbf{Ground-truth} & \textbf{OmniVoice}
& \textbf{Language} & \textbf{Ground-truth} & \textbf{OmniVoice} \\
\hline
\endfirsthead
\hline
\textbf{Language} & \textbf{Ground-truth} & \textbf{OmniVoice}
& \textbf{Language} & \textbf{Ground-truth} & \textbf{OmniVoice} \\
\hline
\endhead
\hline
\endfoot
\hline
\endlastfoot
Afrikaans & 3.54 & 3.08 & Amharic & 4.98 & 7.30 \\
Armenian & 1.38 & 1.45 & Assamese & 5.29 & 3.92 \\
Asturian & 3.79 & 2.07 & Azerbaijani & 2.60 & 3.20 \\
Belarusian & 1.80 & 1.00 & Bengali & 4.10 & 3.08 \\
Bosnian & 1.61 & 0.66 & Bulgarian & 1.66 & 0.87 \\
Burmese & 6.77 & 12.20 & Cantonese & 11.01 & 21.92 \\
Catalan & 1.17 & 0.86 & Cebuano & 1.74 & 1.81 \\
Central Kurdish & 3.92 & 2.97 & Chichewa & 4.63 & 3.32 \\
Chinese & 5.54 & 3.16 & Croatian & 4.64 & 0.93 \\
Czech & 2.19 & 1.21 & Danish & 2.86 & 1.79 \\
Dutch & 2.01 & 1.87 & English & 2.63 & 1.67 \\
Estonian & 0.97 & 1.03 & Filipino & 2.34 & 1.24 \\
Finnish & 1.60 & 1.46 & French & 2.13 & 1.95 \\
Fulah & 12.33 & 7.70 & Galician & 1.52 & 1.03 \\
Ganda & 7.87 & 6.60 & Georgian & 1.78 & 1.83 \\
German & 2.22 & 1.35 & Greek & 2.31 & 1.57 \\
Gujarati & 3.06 & 2.74 & Hausa & 5.14 & 3.03 \\
Hebrew & 4.91 & 9.75 & Hindi & 2.54 & 2.64 \\
Hungarian & 1.75 & 1.04 & Icelandic & 2.82 & 3.77 \\
Igbo & 7.75 & 7.40 & Indonesian & 1.87 & 2.95 \\
Irish & 17.56 & 8.52 & Italian & 1.27 & 1.56 \\
Japanese & 7.70 & 5.96 & Javanese & 3.17 & 2.80 \\
Kabuverdianu & 3.29 & 2.10 & Kamba & 8.81 & 10.42 \\
Kannada & 2.84 & 2.70 & Kazakh & 2.01 & 3.11 \\
Khmer & 6.20 & 11.48 & Kirghiz & 2.33 & 1.80 \\
Korean & 3.72 & 3.78 & Lao & 22.87 & 25.51 \\
Latvian & 1.87 & 1.55 & Lingala & 2.36 & 1.68 \\
Lithuanian & 2.98 & 2.05 & Luo & 3.58 & 3.37 \\
Luxembourgish & 5.48 & 4.52 & Macedonian & 1.05 & 0.89 \\
Malay & 1.59 & 1.19 & Malayalam & 2.99 & 3.40 \\
Maltese & 2.01 & 3.65 & Maori & 4.90 & 1.98 \\
Marathi & 5.02 & 2.76 & Mongolian & 4.78 & 4.26 \\
Nepali & 4.76 & 3.43 & Norwegian Bokmål & 1.50 & 1.14 \\
Occitan & 9.63 & 6.99 & Odia & 5.77 & 5.53 \\
Oromo & 12.50 & 4.97 & Panjabi & 5.76 & 2.95 \\
Pedi & 4.88 & 4.44 & Persian & 1.57 & 1.90 \\
Polish & 1.14 & 0.61 & Portuguese & 1.45 & 1.22 \\
Pushto & 11.42 & 5.53 & Romanian & 2.15 & 0.94 \\
Russian & 1.68 & 1.10 & Serbian & 1.36 & 0.82 \\
Shona & 4.07 & 1.80 & Sindhi & 5.51 & 3.26 \\
Slovak & 2.76 & 0.85 & Slovenian & 2.40 & 1.20 \\
Somali & 10.13 & 4.86 & Spanish & 1.08 & 0.77 \\
Standard Arabic & 2.58 & 1.92 & Swahili & 2.41 & 1.37 \\
Swedish & 2.14 & 1.43 & Tajik & 3.17 & 2.36 \\
Tamil & 4.00 & 3.77 & Telugu & 4.51 & 3.77 \\
Thai & 6.98 & 7.71 & Turkish & 1.94 & 2.71 \\
Ukrainian & 1.45 & 1.23 & Umbundu & 6.97 & 5.44 \\
Urdu & 80.27 & 28.73 & Uzbek & 3.31 & 2.23 \\
Vietnamese & 3.49 & 2.63 & Welsh & 3.91 & 3.37 \\
Wolof & 12.17 & 6.87 & Xhosa & 3.62 & 3.83 \\
Yoruba & 17.97 & 21.37 & Zulu & 3.33 & 2.03
\end{longtable}
\normalsize

\section{Supported Languages of OmniVoice}
OmniVoice supports \textbf{646 languages} with a total of \textbf{581k hours} of training data. Detailed language, along with its OmniVoice language ID, ISO 639-3 code, and training data duration are shown in \autoref{tab:supported_language}.

\tiny
\setlength{\tabcolsep}{1pt}
\renewcommand{\arraystretch}{0.85}

\begin{longtable}{cccc|cccc|cccc}
\caption{Supported languages of OmniVoice along with its OmniVoice language ID, ISO 639-3 code, and training data duration (hours).}
\label{tab:supported_language} \\
\hline
\textbf{Language} & \textbf{ID} & \textbf{ISO} & \textbf{Hours}
& \textbf{Language} & \textbf{ID} & \textbf{ISO} & \textbf{Hours}
& \textbf{Language} & \textbf{ID} & \textbf{ISO} & \textbf{Hours} \\
\hline
\endfirsthead
\hline
\textbf{Language} & \textbf{ID} & \textbf{ISO} & \textbf{Hours}
& \textbf{Language} & \textbf{ID} & \textbf{ISO} & \textbf{Hours}
& \textbf{Language} & \textbf{ID} & \textbf{ISO} & \textbf{Hours} \\
\hline
\endhead
\hline
\endfoot
\hline
\endlastfoot

Abadi & kbt & kbt & 9.73
& Abkhazian & ab & abk & 57.27
& Abron & abr & abr & 9.22 \\
Abua & abn & abn & 10.27
& Adamawa Fulfulde & fub & fub & 13.12
& Adyghe & ady & ady & 32.6 \\
Afade & aal & aal & 10.19
& Afrikaans & af & afr & 4.4
& Agwagwune & yay & yay & 8.26 \\
Aja (Benin) & ajg & ajg & 5.63
& Akebu & keu & keu & 9.1
& Alago & ala & ala & 11.04 \\
Albanian & sq & sqi & 8.59
& Algerian Arabic & arq & arq & 9.64
& Algerian Saharan Arabic & aao & aao & 2.02 \\
Ambo-Pasco Quechua & qva & qva & 9.59
& Ambonese Malay & abs & abs & 10.03
& Amdo Tibetan & adx & adx & 56.94 \\
Amharic & am & amh & 12.83
& Anaang & anw & anw & 9.65
& Angika & anp & anp & 10.65 \\
Antankarana Malagasy & xmv & xmv & 17.9
& Aragonese & an & arg & 16.4
& Arb\"{e}resh\"{e} Albanian & aae & aae & 6.11 \\
Arequipa-La Uni\'{o}n Quechua & qxu & qxu & 10.12
& Armenian & hy & hye & 42.15
& Ashe & ahs & ahs & 10.62 \\
Ash\'{e}ninka Peren\"{e} & prq & prq & 7.16
& Askopan & eiv & eiv & 10.44
& Assamese & as & asm & 270.85 \\
Asturian & ast & ast & 8.48
& Atayal & tay & tay & 7.02
& Awak & awo & awo & 10.22 \\
Ayacucho Quechua & quy & quy & 0.05
& Azerbaijani & az & aze & 9.84
& Baatonum & bba & bba & 10.53 \\
Bacama & bcy & bcy & 9.94
& Bade & bde & bde & 9.89
& Bafia & ksf & ksf & 16.43 \\
Bafut & bfd & bfd & 9.03
& Bagirmi Fulfulde & fui & fui & 15.04
& Bago-Kusuntu & bqg & bqg & 8.86 \\
Baharna Arabic & abv & abv & 10.41
& Bakoko & bkh & bkh & 6.0
& Balanta-Ganja & bjt & bjt & 9.41 \\
Balti & bft & bft & 16.28
& Bamenyam & bce & bce & 9.9
& Bamun & bax & bax & 10.24 \\
Bangwinji & bsj & bsj & 10.0
& Banjar & bjn & bjn & 11.68
& Bankon & abb & abb & 11.2 \\
Baoul\'{e} & bci & bci & 10.21
& Bara Malagasy & bhr & bhr & 12.14
& Barok & bjk & bjk & 10.16 \\
Basa (Cameroon) & bas & bas & 10.66
& Basa (Nigeria) & bzw & bzw & 10.27
& Bashkir & ba & bak & 249.1 \\
Basque & eu & eus & 479.86
& Batak Mandailing & btm & btm & 11.09
& Batanga & bnm & bnm & 15.01 \\
Bateri & btv & btv & 9.8
& Bats & bbl & bbl & 11.22
& Bayot & bda & bda & 9.47 \\
Bebele & beb & beb & 7.52
& Belarusian & be & bel & 1809.43
& Bengali & bn & ben & 271.76 \\
Betawi & bew & bew & 11.15
& Bhili & bhb & bhb & 9.98
& Bhojpuri & bho & bho & 10.05 \\
Bilur & bxf & bxf & 10.84
& Bima & bhp & bhp & 10.67
& Bodo & brx & brx & 231.57 \\
Boghom & bux & bux & 10.48
& Bokyi & bky & bky & 9.85
& Bomu & bmq & bmq & 10.68 \\
Bondei & bou & bou & 9.98
& Borgu Fulfulde & fue & fue & 20.1
& Bosnian & bs & bos & 690.73 \\
Brahui & brh & brh & 19.89
& Braj & bra & bra & 10.68
& Breton & br & bre & 25.48 \\
Buduma & bdm & bdm & 10.17
& Buginese & bug & bug & 11.09
& Bukharic & bhh & bhh & 11.38 \\
Bulgarian & bg & bul & 2190.76
& Bulu (Cameroon) & bum & bum & 9.06
& Bundeli & bns & bns & 10.88 \\
Bunun & bnn & bnn & 9.26
& Bura-Pabir & bwr & bwr & 10.4
& Burak & bys & bys & 9.92 \\
Burmese & my & mya & 12.14
& Burushaski & bsk & bsk & 9.14
& Cacaloxtepec Mixtec & miu & miu & 9.18 \\
Cajatambo North Lima Quechua & qvl & qvl & 9.95
& Cakfem-Mushere & cky & cky & 8.96
& Cameroon Pidgin & wes & wes & 10.06 \\
Campidanese Sardinian & sro & sro & 10.16
& Cantonese & yue & yue & 13302.38
& Catalan & ca & cat & 3358.6 \\
Cebuano & ceb & ceb & 12.17
& Cen & cen & cen & 9.85
& Central Kurdish & ckb & ckb & 137.52 \\
Central Nahuatl & nhn & nhn & 9.51
& Central Pame & pbs & pbs & 9.69
& Central Pashto & pst & pst & 11.4 \\
Central Puebla Nahuatl & ncx & ncx & 9.86
& Central Tarahumara & tar & tar & 9.73
& Central Yupik & esu & esu & 2.18 \\
Central-Eastern Niger Fulfulde & fuq & fuq & 9.28
& Chadian Arabic & shu & shu & 2.29
& Chichewa & ny & nya & 10.8 \\
Chichicapan Zapotec & zpv & zpv & 9.85
& Chiga & cgg & cgg & 10.84
& Chimalapa Zoque & zoh & zoh & 9.35 \\
Chimborazo Highland Quichua & qug & qug & 10.12
& Chinese & zh & cmn & 111343.3
& Chiqui\'{a}n Ancash Quechua & qxa & qxa & 9.99 \\
Chitwania Tharu & the & the & 10.06
& Chokwe & cjk & cjk & 11.01
& Chuvash & cv & chv & 23.96 \\
Cibak & ckl & ckl & 10.91
& Coastal Konjo & kjc & kjc & 10.18
& Copainal\'{a} Zoque & zoc & zoc & 10.07 \\
Cornish & kw & cor & 12.15
& Corongo Ancash Quechua & qwa & qwa & 9.72
& Croatian & hr & hrv & 2795.31 \\
Cross River Mbembe & mfn & mfn & 10.03
& Cuyamecalco Mixtec & xtu & xtu & 9.4
& Czech & cs & ces & 148.13 \\
Dadiya & dbd & dbd & 9.61
& Dagbani & dag & dag & 10.14
& Dameli & dml & dml & 9.18 \\
Danish & da & dan & 1665.98
& Dargwa & dar & dar & 1.22
& Dazaga & dzg & dzg & 9.96 \\
Deccan & dcc & dcc & 10.38
& Degema & deg & deg & 11.07
& Dera (Nigeria) & kna & kna & 11.91 \\
Dghwede & dgh & dgh & 9.95
& Dhatki & mki & mki & 8.83
& Dhivehi & dv & div & 38.61 \\
Dhofari Arabic & adf & adf & 0.31
& Dijim-Bwilim & cfa & cfa & 10.32
& Dogri & dgo & dgo & 117.04 \\
Domaaki & dmk & dmk & 6.38
& Dotyali & dty & dty & 10.85
& Duala & dua & dua & 12.13 \\
Dutch & nl & nld & 2264.13
& D\~{u}ya & ldb & ldb & 11.31
& Dyula & dyu & dyu & 0.34 \\
Eastern Balochi & bgp & bgp & 10.98
& Eastern Bolivian Guaran\'{i} & gui & gui & 22.72
& Eastern Egyptian Bedawi Arabic & avl & avl & 1.86 \\
Eastern Krahn & kqo & kqo & 9.28
& Eastern Mari & mhr & mhr & 272.31
& Eastern Yiddish & ydd & ydd & 18.43 \\
Ebri\'{e} & ebr & ebr & 1.5
& Eggon & ego & ego & 9.95
& Egyptian Arabic & arz & arz & 23.23 \\
Ejagham & etu & etu & 10.3
& Eleme & elm & elm & 11.27
& Eloyi & afo & afo & 11.21 \\
Embu & ebu & ebu & 9.81
& English & en & eng & 206061.1
& Erzya & myv & myv & 3.1 \\
Esan & ish & ish & 10.05
& Esperanto & eo & epo & 1396.64
& Estonian & et & est & 960.37 \\
Eton (Cameroon) & eto & eto & 7.43
& Ewondo & ewo & ewo & 12.71
& Extremaduran & ext & ext & 13.59 \\
Fang (Equatorial Guinea) & fan & fan & 3.51
& Fanti & fat & fat & 11.38
& Farefare & gur & gur & 9.24 \\
Fe'fe' & fmp & fmp & 9.86
& Filipino & fil & fil & 7.71
& Filomena Mata-Coahuitl\'{a}n Totonac & tlp & tlp & 11.35 \\
Finnish & fi & fin & 468.62
& Fipa & fip & fip & 10.55
& French & fr & fra & 23675.32 \\
Fulah & ff & ful & 13.84
& Galician & gl & glg & 208.81
& Gambian Wolof & wof & wof & 9.46 \\
Ganda & lg & lug & 447.82
& Garhwali & gbm & gbm & 19.14
& Gawar-Bati & gwt & gwt & 12.16 \\
Gawri & gwc & gwc & 10.83
& Gbagyi & gbr & gbr & 12.12
& Gbari & gby & gby & 12.59 \\
Geji & gyz & gyz & 10.49
& Gen & gej & gej & 5.39
& Georgian & ka & kat & 156.96 \\
German & de & deu & 21927.13
& Geser-Gorom & ges & ges & 10.08
& Gheg Albanian & aln & aln & 3.92 \\
Ghom\'{a}l\'{a}' & bbj & bbj & 7.32
& Gidar & gid & gid & 10.06
& Glavda & glw & glw & 10.51 \\
Goan Konkani & gom & gom & 9.82
& Goaria & gig & gig & 9.41
& Goemai & ank & ank & 10.0 \\
Gola & gol & gol & 9.26
& Greek & el & ell & 2412.54
& Guarani & gn & grn & 4.06 \\
Guduf-Gava & gdf & gdf & 12.21
& Guerrero Amuzgo & amu & amu & 10.1
& Gujarati & gu & guj & 91.18 \\
Gujari & gju & gju & 8.66
& Gulf Arabic & afb & afb & 98.55
& Gurgula & ggg & ggg & 7.12 \\
Gusii & guz & guz & 9.5
& Gusilay & gsl & gsl & 10.0
& Gweno & gwe & gwe & 8.87 \\
G\"{u}il\'{a} Zapotec & ztu & ztu & 9.17
& Hadothi & hoj & hoj & 10.08
& Hahon & hah & hah & 9.64 \\
Haitian & ht & hat & 0.04
& Hakha Chin & cnh & cnh & 2.24
& Hak\"{o} & hao & hao & 8.56 \\
Halia & hla & hla & 9.86
& Hausa & ha & hau & 17.75
& Hawaiian & haw & haw & 11.79 \\
Hazaragi & haz & haz & 9.69
& Hebrew & he & heb & 13.4
& Hemba & hem & hem & 9.53 \\
Herero & hz & her & 9.59
& Highland Konjo & kjk & kjk & 10.21
& Hijazi Arabic & acw & acw & 22.32 \\
Hindi & hi & hin & 117.17
& Huarijio & var & var & 9.28
& Huautla Mazatec & mau & mau & 6.39 \\
Huaxcaleca Nahuatl & nhq & nhq & 5.07
& Huba & hbb & hbb & 10.7
& Huitepec Mixtec & mxs & mxs & 9.64 \\
Hula & hul & hul & 10.33
& Hungarian & hu & hun & 255.83
& Hunjara-Kaina Ke & hkk & hkk & 8.69 \\
Hwana & hwo & hwo & 11.23
& Ibibio & ibb & ibb & 7.38
& Icelandic & is & isl & 647.29 \\
Idakho-Isukha-Tiriki & ida & ida & 9.31
& Idoma & idu & idu & 11.16
& Igbo & ig & ibo & 13.69 \\
Igo & ahl & ahl & 9.22
& Ikposo & kpo & kpo & 7.83
& Ikwere & ikw & ikw & 10.0 \\
Imbabura Highland Quichua & qvi & qvi & 11.0
& Indonesian & id & ind & 6327.87
& Indus Kohistani & mvy & mvy & 21.64 \\
Interlingua & ia & ina & 13.48
& Inupiaq & ik & ipk & 2.11
& Irish & ga & gle & 21.4 \\
Iron Ossetic & os & oss & 1.38
& Isekiri & its & its & 11.85
& Isoko & iso & iso & 10.33 \\
Italian & it & ita & 9402.46
& Ito & itw & itw & 9.19
& Itz\'{a} & itz & itz & 7.08 \\
Ixtayutla Mixtec & vmj & vmj & 10.17
& Izon & ijc & ijc & 9.95
& Jambi Malay & jax & jax & 10.29 \\
Japanese & ja & jpn & 36914.4
& Jaqaru & jqr & jqr & 9.32
& Jauja Wanca Quechua & qxw & qxw & 11.42 \\
Jaunsari & jns & jns & 11.25
& Javanese & jv & jav & 11.19
& Jiba & juo & juo & 10.43 \\
Jju & kaj & kaj & 10.16
& Judeo-Moroccan Arabic & aju & aju & 7.21
& Juxtlahuaca Mixtec & vmc & vmc & 9.43 \\
Kabardian & kbd & kbd & 108.35
& Kabras & lkb & lkb & 9.99
& Kabuverdianu & kea & kea & 10.51 \\
Kabyle & kab & kab & 529.52
& Kachi Koli & gjk & gjk & 20.83
& Kairak & ckr & ckr & 10.51 \\
Kalabari & ijn & ijn & 11.04
& Kalasha & kls & kls & 9.11
& Kalenjin & kln & kln & 40.42 \\
Kalkoti & xka & xka & 8.0
& Kamba & kam & kam & 14.72
& Kamo & kcq & kcq & 10.49 \\
Kanauji & bjj & bjj & 11.01
& Kanembu & kbl & kbl & 10.19
& Kannada & kn & kan & 128.06 \\
Karekare & kai & kai & 10.52
& Kashmiri & ks & kas & 110.42
& Kathoriya Tharu & tkt & tkt & 10.64 \\
Kati & bsh & bsh & 8.77
& Kazakh & kk & kaz & 1537.29
& Keiyo & eyo & eyo & 9.24 \\
Khams Tibetan & khg & khg & 6.38
& Khana & ogo & ogo & 10.51
& Khetrani & xhe & xhe & 9.4 \\
Khmer & km & khm & 7.1
& Khowar & khw & khw & 15.55
& Kinga & zga & zga & 9.5 \\
Kinnauri & kfk & kfk & 10.32
& Kinyarwanda & rw & kin & 2021.66
& Kirghiz & ky & kir & 46.63 \\
Kirya-Konz\"{a}l & fkk & fkk & 9.98
& Kochila Tharu & thq & thq & 10.28
& Kohistani Shina & plk & plk & 12.75 \\
Kohumono & bcs & bcs & 10.45
& Kok Borok & trp & trp & 10.74
& Kol (Papua New Guinea) & kol & kol & 9.95 \\
Kom (Cameroon) & bkm & bkm & 10.76
& Koma & kmy & kmy & 10.28
& Konkani & knn & knn & 112.83 \\
Konzo & koo & koo & 13.23
& Korean & ko & kor & 8609.28
& Korwa & kfp & kfp & 11.87 \\
Kota (India) & kfe & kfe & 10.25
& Koti & eko & eko & 8.15
& Kuanua & ksd & ksd & 9.91 \\
Kuanyama & kj & kua & 9.88
& Kui (India) & uki & uki & 10.77
& Kulung (Nigeria) & bbu & bbu & 10.39 \\
Kuot & kto & kto & 9.77
& Kushi & kuh & kuh & 10.35
& Kwambi & kwm & kwm & 9.9 \\
Kwasio & nmg & nmg & 10.39
& Lala-Roba & lla & lla & 10.38
& Lamang & hia & hia & 11.07 \\
Lao & lo & lao & 7.63
& Larike-Wakasihu & alo & alo & 9.97
& Lasi & lss & lss & 6.53 \\
Latgalian & ltg & ltg & 27.23
& Latvian & lv & lav & 1441.58
& Levantine Arabic & apc & apc & 15.65 \\
Liana-Seti & ste & ste & 10.43
& Liberia Kpelle & xpe & xpe & 9.5
& Liberian English & lir & lir & 10.26 \\
Libyan Arabic & ayl & ayl & 20.13
& Ligurian & lij & lij & 15.97
& Lijili & mgi & mgi & 10.89 \\
Lingala & ln & lin & 17.99
& Lithuanian & lt & lit & 2629.45
& Loarki & lrk & lrk & 10.5 \\
Logooli & rag & rag & 9.39
& Logudorese Sardinian & src & src & 10.67
& Loja Highland Quichua & qvj & qvj & 10.59 \\
Loloda & loa & loa & 9.31
& Longuda & lnu & lnu & 10.46
& Loxicha Zapotec & ztp & ztp & 9.62 \\
Luba-Lulua & lua & lua & 8.47
& Luo & luo & luo & 36.17
& Lushai & lus & lus & 20.24 \\
Luxembourgish & lb & ltz & 8.46
& Maasina Fulfulde & ffm & ffm & 10.46
& Maba (Chad) & mde & mde & 9.5 \\
Macedo-Romanian & rup & rup & 0.02
& Macedonian & mk & mkd & 27.21
& Mada (Cameroon) & mxu & mxu & 12.0 \\
Mafa & maf & maf & 9.97
& Maithili & mai & mai & 131.37
& Malay & ms & msa & 9.57 \\
Malayalam & ml & mal & 166.57
& Mali & gcc & gcc & 9.87
& Malinaltepec Me'phaa & tcf & tcf & 9.04 \\
Maltese & mt & mlt & 630.29
& Mandara & tbf & tbf & 10.01
& Mandjak & mfv & mfv & 9.55 \\
Manggarai & mqy & mqy & 10.5
& Manipuri & mni & mni & 44.46
& Mansoanka & msw & msw & 9.32 \\
Manx & gv & glv & 10.07
& Maori & mi & mri & 18.02
& Marathi & mr & mar & 156.71 \\
Marghi Central & mrt & mrt & 10.36
& Marghi South & mfm & mfm & 10.05
& Maria (India) & mrr & mrr & 11.0 \\
Marwari (Pakistan) & mve & mve & 9.96
& Masana & mcn & mcn & 10.09
& Masikoro Malagasy & msh & msh & 14.16 \\
Mats\'{e}s & mcf & mcf & 9.61
& Mazaltepec Zapotec & zpy & zpy & 9.47
& Mazatl\'{a}n Mazatec & vmz & vmz & 9.82 \\
Mazatl\'{a}n Mixe & mzl & mzl & 10.05
& Mbe & mfo & mfo & 10.24
& Mbo (Cameroon) & mbo & mbo & 9.51 \\
Mbum & mdd & mdd & 9.82
& Medumba & byv & byv & 10.95
& Mekeo & mek & mek & 9.18 \\
Meru & mer & mer & 9.89
& Mesopotamian Arabic & acm & acm & 3.78
& Mewari & mtr & mtr & 10.58 \\
Min Nan Chinese & nan & nan & 17.55
& Mingrelian & xmf & xmf & 11.47
& Mitlatongo Mixtec & vmm & vmm & 9.95 \\
Miya & mkf & mkf & 10.16
& Mokpwe & bri & bri & 7.53
& Moksha & mdf & mdf & 0.47 \\
Mom Jango & ver & ver & 10.93
& Mongolian & mn & mon & 269.08
& Moroccan Arabic & ary & ary & 104.67 \\
Motu & meu & meu & 9.88
& Mpiemo & mcx & mcx & 9.88
& Mpumpong & mgg & mgg & 4.94 \\
Mundang & mua & mua & 9.2
& Mungaka & mhk & mhk & 7.53
& Musey & mse & mse & 7.21 \\
Musgu & mug & mug & 4.74
& Musi & mui & mui & 10.52
& Naba & mne & mne & 10.37 \\
Najdi Arabic & ars & ars & 203.54
& Nalik & nal & nal & 10.33
& Nawdm & nmz & nmz & 6.3 \\
Ndonga & ng & ndo & 9.08
& Neapolitan & nap & nap & 9.97
& Nepali & npi & npi & 171.5 \\
Ngamo & nbh & nbh & 10.04
& Ngas & anc & anc & 10.14
& Ngiemboon & nnh & nnh & 16.15 \\
Ngizim & ngi & ngi & 10.06
& Ngomba & jgo & jgo & 10.15
& Ngombale & nla & nla & 8.79 \\
Nigerian Fulfulde & fuv & fuv & 9.97
& Nigerian Pidgin & pcm & pcm & 11.04
& Nimadi & noe & noe & 11.12 \\
Nobiin & fia & fia & 9.96
& North Mesopotamian Arabic & ayp & ayp & 10.92
& North Moluccan Malay & max & max & 9.43 \\
Northern Betsimisaraka Malagasy & bmm & bmm & 19.12
& Northern Hindko & hno & hno & 20.04
& Northern Kurdish & kmr & kmr & 69.59 \\
Northern Pame & pmq & pmq & 10.24
& Northern Pashto & pbu & pbu & 11.03
& Northern Uzbek & uzn & uzn & 15.23 \\
Northwest Gbaya & gya & gya & 8.45
& Norwegian & no & nor & 3849.8
& Norwegian Bokm{\aa}l & nb & nob & 12.7 \\
Norwegian Nynorsk & nn & nno & 1.54
& Notsi & ncf & ncf & 9.84
& Nyankpa & yes & yes & 10.26 \\
Nyungwe & nyu & nyu & 8.98
& Nzanyi & nja & nja & 10.02
& N\"{u}pode Huitoto & hux & hux & 9.04 \\
Occitan & oc & oci & 16.8
& Od & odk & odk & 20.26
& Odia & ory & ory & 144.81 \\
Odual & odu & odu & 10.57
& Omani Arabic & acx & acx & 22.03
& Orizaba Nahuatl & nlv & nlv & 11.42 \\
Orma & orc & orc & 22.01
& Ormuri & oru & oru & 16.74
& Oromo & om & orm & 6.6 \\
Pahari-Potwari & phr & phr & 24.03
& Paiwan & pwn & pwn & 13.76
& Panjabi & pa & pan & 147.37 \\
Papuan Malay & pmy & pmy & 10.17
& Parkari Koli & kvx & kvx & 11.04
& Pedi & nso & nso & 12.64 \\
Pero & pip & pip & 9.85
& Persian & fa & fas & 366.07
& Petats & pex & pex & 10.2 \\
Phalura & phl & phl & 20.69
& Piemontese & pms & pms & 16.01
& Piya-Kwonci & piy & piy & 10.38 \\
Plateau Malagasy & plt & plt & 19.39
& Polish & pl & pol & 911.68
& Poqomam & poc & poc & 9.63 \\
Portuguese & pt & por & 16855.05
& Pulaar & fuc & fuc & 14.77
& Pular & fuf & fuf & 13.77 \\
Puno Quechua & qxp & qxp & 9.81
& Pushto & ps & pus & 88.62
& P\"{o}koot & pko & pko & 10.4 \\
Qaqet & byx & byx & 9.79
& Quiotepec Chinantec & chq & chq & 9.76
& Rana Tharu & thr & thr & 9.99 \\
Rangi & lag & lag & 9.47
& Rapoisi & kyx & kyx & 9.17
& Ratahan & rth & rth & 9.34 \\
Ray\'{o}n Zoque & zor & zor & 9.04
& Romanian & ro & ron & 70.23
& Romansh & rm & roh & 9.21 \\
Rombo & rof & rof & 18.9
& Rotokas & roo & roo & 9.07
& Rukai & dru & dru & 9.26 \\
Russian & ru & rus & 20338.5
& Sacapulteco & quv & quv & 8.9
& Saidi Arabic & aec & aec & 9.28 \\
Sakalava Malagasy & skg & skg & 9.02
& Sakizaya & szy & szy & 11.47
& Saleman & sau & sau & 10.53 \\
Samba Daka & ccg & ccg & 10.11
& Samba Leko & ndi & ndi & 11.27
& San Felipe Otlaltepec Popoloca & pow & pow & 8.84 \\
San Francisco Del Mar Huave & hue & hue & 9.45
& San Juan Atzingo Popoloca & poe & poe & 10.01
& San Mart\'{i}n Itunyoso Triqui & trq & trq & 8.29 \\
San Miguel El Grande Mixtec & mig & mig & 9.66
& Sansi & ssi & ssi & 10.47
& Sanskrit & sa & san & 84.44 \\
Santa Ana de Tusi Pasco Quechua & qxt & qxt & 10.05
& Santa Catarina Albarradas Zapotec & ztn & ztn & 10.02
& Santali & sat & sat & 98.37 \\
Santiago del Estero Quichua & qus & qus & 9.55
& Saposa & sps & sps & 9.81
& Saraiki & skr & skr & 4.13 \\
Sardinian & sc & srd & 2.77
& Saya & say & say & 10.02
& Sediq & trv & trv & 7.77 \\
Serbian & sr & srp & 1855.33
& Seri & sei & sei & 9.81
& Shina & scl & scl & 9.84 \\
Shona & sn & sna & 9.96
& Siar-Lak & sjr & sjr & 9.87
& Sibe & nco & nco & 9.96 \\
Sicilian & scn & scn & 13.35
& Sihuas Ancash Quechua & qws & qws & 10.18
& Sikkimese & sip & sip & 10.07 \\
Sinaugoro & snc & snc & 10.38
& Sindhi & sd & snd & 46.27
& Sindhi Bhil & sbn & sbn & 10.53 \\
Sinhala & si & sin & 11.98
& Sinicahua Mixtec & xti & xti & 9.5
& Sipacapense & qum & qum & 9.37 \\
Siwai & siw & siw & 10.47
& Slovak & sk & slk & 2478.46
& Slovenian & sl & slv & 1172.61 \\
Solos & sol & sol & 9.95
& Somali & so & som & 13.22
& Soninke & snk & snk & 10.04 \\
South Giziga & giz & giz & 10.03
& South Ucayali Ash\'{e}ninka & cpy & cpy & 9.15
& Southeastern Nochixtl\'{a}n Mixtec & mxy & mxy & 9.48 \\
Southern Betsimisaraka Malagasy & bzc & bzc & 17.45
& Southern Pashto & pbt & pbt & 11.6
& Southern Pastaza Quechua & qup & qup & 11.13 \\
Soyaltepec Mazatec & vmp & vmp & 10.17
& Spanish & es & spa & 27559.74
& Standard Arabic & arb & arb & 1483.53 \\
Standard Moroccan Tamazight & zgh & zgh & 1.19
& Sudanese Arabic & apd & apd & 9.93
& Sulka & sua & sua & 10.12 \\
Svan & sva & sva & 15.11
& Swahili & sw & swa & 418.41
& Swedish & sv & swe & 2453.14 \\
Tae' & rob & rob & 9.02
& Tahaggart Tamahaq & thv & thv & 4.25
& Taita & dav & dav & 9.12 \\
Tajik & tg & tgk & 9.23
& Tamil & ta & tam & 423.09
& Tandroy-Mahafaly Malagasy & tdx & tdx & 3.81 \\
Tangale & tan & tan & 10.14
& Tanosy Malagasy & txy & txy & 12.07
& Tarok & yer & yer & 10.08 \\
Tatar & tt & tat & 30.03
& Tedaga & tuq & tuq & 10.0
& Telugu & te & tel & 230.21 \\
Tem & kdh & kdh & 4.07
& Teop & tio & tio & 9.85
& Tepeuxila Cuicatec & cux & cux & 7.83 \\
Tepinapa Chinantec & cte & cte & 9.54
& Tera & ttr & ttr & 9.89
& Terei & buo & buo & 9.48 \\
Termanu & twu & twu & 11.45
& Tesaka Malagasy & tkg & tkg & 17.86
& Tetelcingo Nahuatl & nhg & nhg & 8.92 \\
Teutila Cuicatec & cut & cut & 8.04
& Thai & th & tha & 10499.77
& Tibetan & bo & bod & 82.27 \\
Tida\'{a} Mixtec & mtx & mtx & 9.09
& Tidore & tvo & tvo & 10.31
& Tigak & tgc & tgc & 9.71 \\
Tigre & tig & tig & 7.49
& Tigrinya & ti & tir & 0.08
& Tilquiapan Zapotec & zts & zts & 9.33 \\
Tinputz & tpz & tpz & 9.33
& Tlacoapa Me'phaa & tpl & tpl & 9.28
& Tlacoatzintepec Chinantec & ctl & ctl & 10.04 \\
Tlingit & tli & tli & 0.41
& Toki Pona & tok & tok & 13.51
& Tomoip & tqp & tqp & 10.1 \\
Tondano & tdn & tdn & 9.14
& Tonsea & txs & txs & 9.32
& Tooro & ttj & ttj & 10.31 \\
Torau & ttu & ttu & 9.87
& Torwali & trw & trw & 14.98
& Tsimihety Malagasy & xmw & xmw & 11.53 \\
Tsotso & lto & lto & 9.77
& Tswana & tn & tsn & 4.24
& Tugen & tuy & tuy & 8.79 \\
Tuki & bag & bag & 10.97
& Tula & tul & tul & 9.79
& Tulu & tcy & tcy & 11.72 \\
Tunen & tvu & tvu & 9.85
& Tungag & lcm & lcm & 9.77
& Tunisian Arabic & aeb & aeb & 21.63 \\
Tupuri & tui & tui & 9.26
& Turkana & tuv & tuv & 10.17
& Turkish & tr & tur & 125.36 \\
Turkmen & tk & tuk & 2.86
& Tututepec Mixtec & mtu & mtu & 10.13
& Twi & tw & twi & 0.25 \\
Ubaghara & byc & byc & 11.11
& Uighur & ug & uig & 428.77
& Ukrainian & uk & ukr & 1851.97 \\
Umbundu & umb & umb & 10.59
& Upper Sorbian & hsb & hsb & 2.71
& Urdu & ur & urd & 211.27 \\
Ushojo & ush & ush & 6.36
& Uzbek & uz & uzb & 115.28
& Vai & vai & vai & 8.76 \\
Vietnamese & vi & vie & 8481.98
& Votic & vot & vot & 0.1
& V\"{o}ro & vro & vro & 15.66 \\
Waci Gbe & wci & wci & 8.02
& Wadiyara Koli & kxp & kxp & 20.0
& Waja & wja & wja & 10.22 \\
Wakhi & wbl & wbl & 11.67
& Wanga & lwg & lwg & 9.36
& Wapan & juk & juk & 10.22 \\
Warji & wji & wji & 11.39
& Welsh & cy & cym & 131.21
& Wemale & weo & weo & 9.09 \\
Western Frisian & fy & fry & 70.41
& Western Highland Purepecha & pua & pua & 10.17
& Western Juxtlahuaca Mixtec & jmx & jmx & 10.01 \\
Western Maninkakan & mlq & mlq & 9.83
& Western Mari & mrj & mrj & 32.26
& Western Niger Fulfulde & fuh & fuh & 9.69 \\
Western Panjabi & pnb & pnb & 10.0
& Wolof & wo & wol & 8.71
& Wuzlam & udl & udl & 9.23 \\
Xanagu\'{i}a Zapotec & ztg & ztg & 9.86
& Xhosa & xh & xho & 13.35
& Yace & ekr & ekr & 10.76 \\
Yakut & sah & sah & 16.08
& Yalahatan & jal & jal & 11.18
& Yanahuanca Pasco Quechua & qur & qur & 9.95 \\
Yangben & yav & yav & 8.7
& Yaqui & yaq & yaq & 9.93
& Yauyos Quechua & qux & qux & 9.35 \\
Yekhee & ets & ets & 10.11
& Yiddish & yi & yid & 1.81
& Yidgha & ydg & ydg & 9.89 \\
Yoruba & yo & yor & 15.66
& Yutanduchi Mixtec & mab & mab & 9.26
& Zacatl\'{a}n-Ahuacatl\'{a}n-Tepetzintla Nahuatl & nhi & nhi & 0.05 \\
Zarma & dje & dje & 10.72
& Zaza & zza & zza & 1.52
& Zulu & zu & zul & 14.83 \\
\"{O}mie & aom & aom & 8.19
\\
\end{longtable}
\
\normalsize

\section{Limitations and Future Works}

OmniVoice is currently trained solely on publicly available open-source datasets. Due to inconsistent annotation quality and uneven acoustic quality across these sources, substantial room remains for performance improvement when trained on more curated, high-quality data. Likewise, the scope and flexibility of the model’s instruction-following capacity are constrained by the diversity and quality of existing instruction-tuning data. Constructing high-quality instruction datasets would therefore notably enhance the model’s voice design and customization capabilities.
Furthermore, OmniVoice has not yet been optimized for processing complex numeric sequences or mathematical patterns. Integrating an external text-normalization front-end would effectively strengthen the model’s performance in such scenarios. Additionally, unlike continuous-space NAR TTS models, where inference steps can be drastically reduced via techniques such as flow distillation~\cite{zhu2025zipvoice}, no existing approach enables comparable inference acceleration for discrete-space NAR TTS models. Exploring optimization strategies to decrease inference steps for this branch of models thus constitutes a promising direction for future work.

\section{Ethics Statements}

This work is intended only for academic research purposes. The proposed TTS model is not used for commercial purposes by the authors’ affiliated institutions. Given its ability to synthesize speech with high speaker similarity, it carries potential risks of misuse. Any illegal use of this model is strictly prohibited.

\end{document}